\documentclass[10pt,twocolumn,letterpaper]{article}

\usepackage{cvpr}              

\usepackage[dvipsnames]{xcolor}

\definecolor{my_purple}{HTML}{9903F0}
\definecolor{my_green}{HTML}{156C09}
\definecolor{my_orange}{HTML}{FF3300}

\definecolor{cvprblue}{rgb}{0.21,0.49,0.74}
\usepackage[pagebackref,breaklinks,colorlinks,citecolor=cvprblue]{hyperref}
\usepackage{pifont}
\usepackage[none]{hyphenat}
\usepackage{float}
\usepackage{verbatim}
\usepackage{xcolor}
\definecolor{main_method_blue}{RGB}{148,169,216}
\definecolor{main_method_orange}{RGB}{234,180,138}
\definecolor{hash_fusion_red}{RGB}{192,0,0}
\definecolor{Ld_Lc_green}{RGB}{94,129,63}

\usepackage{amsmath}
\usepackage{amsthm}
\usepackage{eucal}
\usepackage{amssymb}
\usepackage{mathrsfs}
\usepackage{bm}
\usepackage{multirow}

\usepackage[accsupp]{axessibility} 

\usepackage{tikz}

\def\paperID{3216} 
\def\confName{CVPR}
\def\confYear{2024}

\title{How Far Can We Compress Instant-NGP-Based NeRF?}

\author{
  Yihang Chen \textsuperscript{1, 2} \qquad Qianyi Wu \textsuperscript{2} \qquad Mehrtash Harandi \textsuperscript{2} \qquad Jianfei Cai \textsuperscript{2} \\
  \textsuperscript{1}Shanghai Jiao Tong University \qquad
  \textsuperscript{2}Monash University \\
  {\tt\small yhchen.ee@sjtu.edu.cn, \{qianyi.wu, mehrtash.harandi, jianfei.cai\}@monash.edu}
}

\begin{document}
\maketitle
\begin{abstract}

In recent years, Neural Radiance Field (NeRF) has demonstrated remarkable capabilities in representing 3D scenes. 
To expedite the rendering process, learnable explicit representations have been introduced for combination with implicit NeRF representation, which however results in a large storage space requirement. 
%
In this paper, we introduce the Context-based NeRF Compression (CNC) framework, which leverages highly efficient context models to provide a storage-friendly NeRF representation.  
Specifically, we excavate both level-wise and dimension-wise context dependencies to enable probability prediction for information entropy reduction.
Additionally, we exploit hash collision and occupancy grids as strong prior knowledge for better context modeling.
To the best of our knowledge, we are the first to construct and exploit context models for NeRF compression.
We achieve a size reduction
of 100$\times$ and 70$\times$ with improved fidelity against the baseline Instant-NGP on Synthesic-NeRF and Tanks and Temples datasets, respectively. Additionally, we attain 86.7\% and 82.3\% storage size reduction against the SOTA NeRF compression method BiRF. 
%
Our code is available here: \textcolor{red}{https://github.com/YihangChen-ee/CNC}.
\end{abstract}    
\section{Introduction}
\label{sec:intro}
\begin{figure}
    \centering
    \includegraphics[width=1\linewidth]{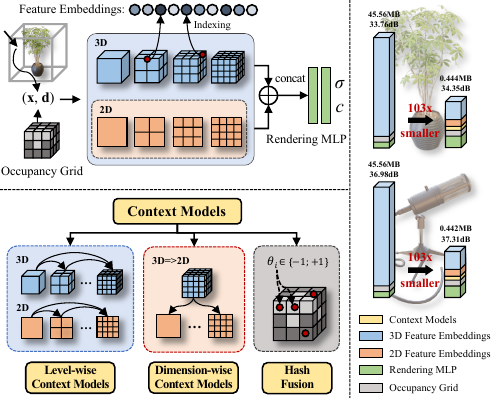}
    \caption{Motivation of our work. Instant-NGP represents 3D scenes using 3D hash feature embeddings along with a rendering MLP, which takes a non-negligible storage size with the embeddings accounting for over 99\% of storage size (upper-left). To tackle this, we introduce context models to substantially compress feature embeddings, with the three key technical components (bottom-left). \emph{Our approach achieves a size reduction of over 100$\times$ while simultaneously improving fidelity.}\protect\footnotemark[1]}
    \vspace{-15pt}
    \label{fig:teaser}
\end{figure}
\footnotetext[1]{The size of INGP is calculated under 16 levels with resolution from 16 to 2048. The feature vector dimension is 2 and represented with FP32.}

High-quality photo-realistic rendering at novel viewpoints remains a pivotal challenge in both computer vision and computer graphics. 
Traditional explicit 3D representations, such as voxel grids~\cite{seitz1999photorealistic,kutulakos2000theory,sitzmann2019deepvoxels,neuralvolume}, have earned their place due to their efficiency across numerous applications. However, their discrete nature makes them susceptible to the limitations imposed by the Nyquist sampling theorem, often necessitating exponentially increased memory for capturing detailed nuances.

In the past few years, Neural Radiance Field (NeRF)~\cite{NeRF} has emerged as a game-changer for novel view synthesis. NeRF defines both density and radiance at a 3D point as functions of the 3D coordinates. Its implicit representation, encapsulated within a Multi-Layer Perceptron (MLP), captures continuous signals of a 3D scene seamlessly. Leveraging frequency-based positional embeddings of 3D coordinates~\cite{NeRF,tancik2020fourier,yang2023freenerf}, NeRF has showcased superior novel view synthesis quality in comparison to traditional explicit 3D representations. While NeRF exhibits good characteristics in memory efficiency and image quality, its complex queries of the MLP slow down its rendering speed.

To boost NeRF's rendering speed, recent approaches have converged towards a hybrid representation, merging explicit voxelized feature encoding with implicit neural networks. This combination promises faster rendering without compromising on quality. These methods include varied data structures such as dense grids~\cite{Plenoxels,DVGO,takikawa2022variable,takikawa2021neural}, octrees~\cite{yu2021plenoctrees,martel2021acorn}, sparse voxel grids~\cite{liu2020neural}, and hash tables~\cite{INGP}. Among them,  Instant-NGP (INGP)~\cite{INGP} which introduces multi-resolution learnable hash embeddings is the most representative one. These hybrid strategies are fast becoming staples in modern NeRF architectures~\cite{barron2023zip,tancik2023nerfstudio}. Yet, with gains in rendering quality and speed, storage is becoming the new constraint. For example, with the occurrence of large-scale NeRFs~\cite{tancik2022block,CityNeRF,li2023matrixcity}, the total storage of their parameters restricts their accessibility and deployment. The storage challenge becomes even more pressing when further considering numerous 3D scenes. 

This leads us to ponder: \emph{Can we reduce the storage cost of modern NeRFs with hybrid representations such as Instant-NGP while maintaining high fidelity and rendering speed?} A few NeRF compression methods have been proposed to address this. 
The common idea is to follow the ``Deep Compression''\cite{deepcompression} concept, which relies on pruning and quantization techniques to squeeze the explicit feature encoding segment. For example,  VQRF~\cite{VQRF} pioneers the trimming of redundant voxel grids and employs vector quantization for parameter reduction. BiRF~\cite{BiRF} goes a step further, using 1-bit binarization for feature embeddings compression. While these methods notably reduce storage needs, we advocate that the efficiency of voxel feature encoding can be further improved from a data compression perspective. Our core principle is to decrease the information uncertainty (entropy) of voxel feature encoding, which has been widely investigated in image and video compression but rarely explored in NeRF compression.  
By leveraging efficient entropy codecs like Arithmetic Coding (AE)~\cite{AE}, we aim to achieve a balance between minimizing storage cost and maintaining rendering quality and speed.

In this paper, we propose a Context-based NeRF Compression (CNC) framework, a pioneering approach to create a storage-optimized NeRF model. Based on Instant-NGP~\cite{INGP} and its multi-resolution hash encoding, our model offers both rendering quality and efficiency. Our core proposition lies in the entropy minimization of explicit feature encoding using accurate context models. Specifically, we introduce a meticulously designed entropy estimation function for each resolution in feature embeddings, on the assumption of Bernoulli distribution. This is coupled with both level-wise and dimension-wise context models that combine different aspects of the hashing embeddings, see Fig.~\ref{fig:teaser}. We also leverage the hash collision and the occupancy grid from Instant-NGP to further ensure our context models' accuracy. 
In summary, the major contributions of this work are threefold:
\begin{enumerate}
    \item 
    To our knowledge, we are the first to propose to model the contexts of 
    INGP's multi-resolution hashing feature embeddings 
    to effectively reduce storage size while maintaining fidelity and speed simultaneously.
    \item 
    We design customized context models that effectively build not only multi-level but also cross-dimension dependencies for INGP hash embeddings. We also utilize hash collision and occupancy grid as strong prior knowledge to provide more accurate contexts.
    \item Extensive experiments show that our CNC framework achieves a size reduction of over 100$\times$ and 70$\times$ while simultaneously improving fidelity, compared to the baseline INGP, on Synthetic-NeRF and Tanks and Temples datasets, respectively. Our approach significantly outperforms the SOTA NeRF compression algorithm, BiRF~\cite{BiRF}, with over 80\% 
    size reduction.
    %
\end{enumerate}

\section{Related work}
\label{sec:relatedwork}

\textbf{Neural radiance field: from implicit to explicit.}
In recent years, Neural Radiance Field (NeRF)~\cite{NeRF} has significantly advanced the area of novel view synthesis by effectively reconstructing 3D radiance fields in a neural implicit way.
Specifically, NeRF utilizes a coordinate-based implicit Multi-Layer Perceptron (MLP) to enable synthesis from arbitrary views. Nevertheless, due to the absence of scene-specific information in the input coordinates, the MLP is designed to be relatively complex to encompass all necessary information. Such complexity slows down 
the entire rendering process, resulting in days for training. 

To expedite rendering, diverse data structures have been introduced as input to explicitly carry scene-specific information, to reduce or even eliminate 
the MLP to achieve much faster rendering. 
For example, Instant-NGP (INGP)~\cite{INGP}, TensoRF~\cite{TensoRF} and K-Planes~\cite{Kplanes} employ learnable embeddings or voxels to represent 3D scenes, which significantly reduce the computational burden of the rendering MLP. Plenoxels~\cite{Plenoxels} and DVGO~\cite{DVGO} take this a step further by eliminating the entire implicit MLP and opting for a purely explicit representation of the whole 3D scene. However, one major downside of these explicit representations is the substantial parameter size, sometimes reaching hundreds of MBs~\cite{DVGO, Plenoxels}, which results in undesirably large storage costs, especially taking into account a vast number of 3D scenes. To address this issue, compression techniques are emerging for more compact NeRF representations. In this paper, we explore context models for the representative INGP-based structure and push NeRF compression to a new level.

\noindent \textbf{Compression techniques: which is the most suitable?}
Before delving into NeRF compression, we would like to start with a glance at existing compression techniques.
First and foremost, model compression stands as a significant category. Given that different model weights exert varying impacts on the final results, various approaches compress them based on weight significance via pruning~\cite{zhu2017prune}, quantization~\cite{model_quantization}, and low-rank approximation~\cite{low-rank-1, low-rank-2}. Knowledge distillation~\cite{distillation} is another avenue in which student models are guided by teachers to create much more compact versions. 
With the evaluated importance of parameters in NeRF models, some NeRF compression algorithms
select the most representative ones to retain information using codebooks~\cite{VQRF, CompactNeRF} or gradients~\cite{Re:NeRF}. Among the existing NeRF compression algorithms, BiRF~\cite{BiRF} achieves SOTA Rate-Distortion (RD) performance by utilizing quantization techniques to binarize hash embeddings of INGP-based NeRF. 

Apart from leveraging weight importance, contextual dependencies among neighboring elements are another essential source for compression, which has been widely exploited as spatial relations in image compression~\cite{cheng2020learned, he2021checkerboard, he2022elic}, and as both spatial and temporal relations in video compression~\cite{DCVC, DCVC-TCM, DCVC-HEM}. Some recent NeRF compression methods also exploit spatial relations by utilizing    
%
techniques such as rank-residual decomposition~\cite{CCNeRF}, wavelet decomposition~\cite{MaskDWT}, or probability models~\cite{SHACIRA} to achieve better compression. However, all these approaches often overlook the unique structures of NeRFs, failing to fully extract contextual information. In contrast, our work discovers that the multi-level embeddings in INGP-based NeRFs exhibit highly organized structures, and introduces efficient context models to effectively model contextual relations at different levels and dimensions, 
which leads to remarkable improvement in rate-distortion (RD) performance. 

\begin{figure*}
    \centering
    \includegraphics[width=0.95\linewidth]{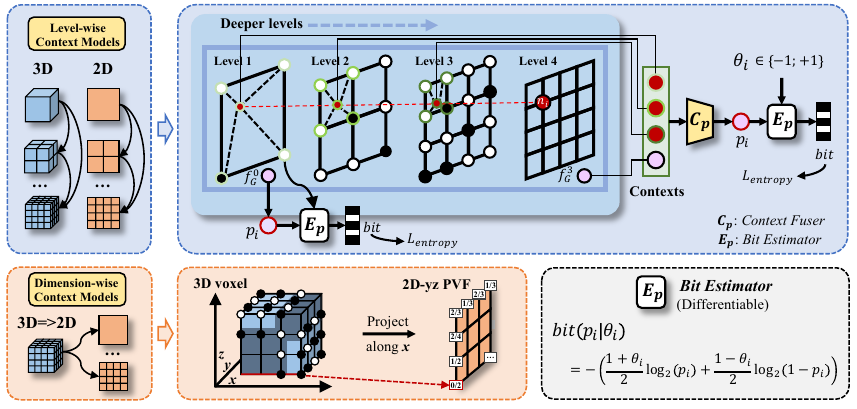}
    \caption{Overview of the proposed level-wise and dimension-wise context models. In the level-wise context model \textcolor{main_method_blue}{(dashed blue box)},  we first find the vertex $n_i$ of the feature vector $\theta_i$ using hash function and then estimate its distribution probability $p_i$ by a \emph{Context Fuser} $\bm{C_p}$ with aggregated contexts from previously decoded levels. It's worth noting that while the illustration here is 2D, the same approach applies to 3D using trilinear interpolation. In the dimension-wise context models \textcolor{main_method_orange}{(dashed orange box)}, the last level of 3D voxel is projected onto 2D planes to obtain Projected Voxel Feature (PVF), which is then used for context interpolation. Deep-blue areas on the voxels indicate empty cells of the occupancy grid. At bottom-right \textcolor{black}{(dashed black box)}, the formula of the entropy-based \emph{Bit Estimator} $\bm{E_p}$ is provided, which is carefully designed to ensure a more efficient backward gradient.}
    \vspace{-15pt}
    \label{fig:main_method}
\end{figure*}

\section{Method}\label{sec:method}
Our objective is to develop a \emph{storage-friendly} NeRF with efficient rendering speed and high fidelity. Our approach builds upon Instant-NGP (INGP)~\cite{INGP}. As shown in the right of Fig.~\ref{fig:teaser}, the primary storage of INGP comes from explicit hash feature embeddings. To minimize the overall model size, we introduce a novel framework named Context-based NeRF Compression (CNC), comprising various modules as depicted in Fig.~\ref{fig:main_method}. The technical details are elaborated in the following subsections.

\subsection{Preliminaries}\label{sec:preliminaries}
\noindent\textbf{Neural Radiance Field}~\cite{NeRF} renders a 3D scene through an implicit rendering MLP. This MLP, when provided with the input coordinate $\mathbf{x}\in\mathbb{R}^3$ and viewing direction $\mathbf{d}\in\mathbb{R}^2$, can generate density 
$\sigma(\mathbf{x})$ and color $\mathbf{c}(\mathbf{x}, \mathbf{d})$ for rendering. Given a ray $\mathbf{r}(v)=\mathbf{o} + v\mathbf{d}$ casting from the camera $\mathbf{o}\in\mathbb{R}^3$, the rendered pixel color $\hat{C}$ can be calculated by accumulating the density and color along the ray~\cite{volume_rendering}, \ie:
\begin{equation}
    \hat{C}(\mathbf{r}) =\int_{v_n}^{v_f}T(v)\sigma(\mathbf{r}(v))\mathbf{c}\big(\mathbf{r}(v), \mathbf{d}\big)dv, \label{eq:volume_rendering}
\end{equation}
where $\sigma(\mathbf{r}(v))$ is the density at the sampled point and $T(v)=\exp\Big(-\int_{v_n}^v\sigma(\mathbf{r}(u))du\Big)$ measures the transmittance along the ray. 
To enhance the representation of high-frequency details, NeRF proposes to map the input coordinates with a frequency-based position encoding~\cite{NeRF}. However, the extensive querying of the heavy MLP slows down the training and inference processes. 

\noindent\textbf{Instant-NGP}~\cite{INGP}. To expedite the rendering process of NeRF, INGP~\cite{INGP} introduces the concept of multi-level feature embeddings as a novel approach to positional encoding, where deeper levels correspond to voxels with higher resolutions. This allows for the utilization of a more compact rendering MLP without compromising the quality. For a given 3D coordinate $\mathbf{x}$, it is situated within a voxel at each level. For each resolution level $l\in \{1, \dots, L\}$, the feature at $\mathbf{x}$ can be calculated by interpolating from the features on the vertex features in the surrounding voxel grid, \ie $\mathbf{f}^l(\mathbf{x})=\rm{interp}(\mathbf{x},\Theta)$, where $\Theta = \{\bm{\theta}_i^l=(\theta_i^{l,1}, \dots, \theta_i^{l,F})\in \mathbb{R}^F| i=1,\dots, T^l\}$ is the trainable feature embedding collection, $F$ is the dimension of each feature vector $\bm{\theta}_i^l$, and $T^l$ is the size of the feature embedding set $\Theta$. For each level, when the resolution of the voxel grid exceeds a specified threshold, the vertex features will be acquired through a spatial hashing function~\cite{teschner2003optimized} to query $\Theta$ for efficiency. The interpolated features from different levels are then concatenated together and fed into the size-reduced rendering MLP for reconstruction. Another technique that INGP employs to accelerate rendering is the occupancy grid, which skips the empty space by efficient ray sampling. More details can be found in~\cite{INGP}. Consequently, the total storage of INGP includes the feature embeddings, the occupancy grid and the rendering MLP, as shown in Fig.~\ref{fig:teaser}.

\noindent\textbf{BiRF}~\cite{BiRF}. While the use of implicit feature embeddings significantly enhances rendering speed, it concurrently imposes a storage burden. The state-of-the-art method BiRF~\cite{BiRF} introduces an innovative approach by binarizing $\theta$ in feature embeddings to $\{-1, +1\}$ using a $\rm{sign}$ function and backpropagating them through a straight-through estimator~\cite{ste}.  This quantization solution reduces the model size by a large margin. Additionally, BiRF shows that introducing extra tri-plane features can enhance reconstruction quality with a similar number of parameters. In this work, we follow their model design with hybrid 2D-3D feature embeddings for the radiance field and build our context models on top of that.

\subsection{Compress Embeddings with Context Model}
Without loss of generality, we omit the notation of resolution level $l$ from $\bm{\theta}_i^l$
and assume the feature dimension $F$ is 1 for simplicity, for which $\bm{\theta_i}=\theta_i$. 
The fundamental principle of our framework is to decrease the information uncertainty of $\theta_i$. Inspired from the binarization concept of BiRF~\cite{BiRF}, we model each value $\theta_i$ to conform to a Bernoulli distribution, \ie $\theta_i \in \{-1, +1\}$. This results in a differentiable bit consumption estimator, based on entropy, for each $\theta_i$
with the probability $p_i = \mathbb{P}(\theta_i=+1)\in[0, 1]$:
\begin{equation}
\label{eq:inforamtion_entropy}
\begin{aligned}
    \textit{bit}(p_i|\theta_i)
    =& -\Big(\frac{1+\theta_i}{2}\log_2(p_i)+\frac{1-\theta_i}{2}\log_2(1-p_i)\Big) \\
    =& \left\{
        \begin{array}{ll}
            -\log_2(p_i) & \quad \theta_i = +1 \\
            -\log_2(1-p_i) & \quad \theta_i = -1
        \end{array}
    \right.
\end{aligned}
\end{equation}
A straightforward method to estimate $p_i$ is to use the occurrence frequency 
$f_{G} = \frac{\#\{\theta_i| \theta_i=+1, \theta_i\in\Theta\}}{\#\{\theta_i| \theta_i\in\Theta\}}$, where $\#$ denotes the number counting, 
such that $p_i=f_{G}$ for $i=1,\dots, T$. However, we find this manner is suboptimal as $f_{G}$ is not accurate for all the embeddings. Our key insight is that the spatial context in 3D space can enhance the precision of $p_i$ estimation. For instance, if a point is empty in 3D space, we should spend fewer bits to store the corresponding features in $\Theta$. This motivates us to introduce context models in the spatial domain when estimating $p_i$. Particularly, we propose two types of context models: level-wise and dimension-wise.

\subsection{Level-Wise Context Models}
\label{subsec:level_wise_context_models}
The primary goal of the level-wise context models is to establish contextual dependencies among $\theta_i$s across different levels, with the expectation that more accurately predicted probability $p_i$s lead to size reduction. Several critical issues need to be taken into consideration:
\begin{enumerate}
    \item Contextual dependencies should obey causal processes. That is we can only utilize $\theta_i$s that have already been decoded as contexts to predict those yet to be decoded.
    \item Context models themselves also consume storage space. This limitation prevents us from adopting arbitrarily large context models, even though having more parameters could enhance their prediction.
    \item The order of contextual dependencies is of great importance. If more informative parts are decoded first, they can provide more context to others but at the cost of consuming more bits to store themselves. 
\end{enumerate}

In light of these considerations, we have designed our level-wise context models in a coarse-to-fine manner, as illustrated in the \textcolor{main_method_blue}{dashed blue box} of Fig.~\ref{fig:main_method} (upper). 
Consider an example of vertex $n_i$ with associated feature $\theta_i$ at the current level $l=4$, as shown in Fig.~\ref{fig:main_method}.  
Following the coarse-to-fine principle, the context of $n_i$ depends on the interpolated features at the corresponding location from the previous $L_c$ levels, where $L_c$ is a preset constant (\eg, $L_c=3$ in Fig.~\ref{fig:main_method}). We also incorporate the occurrence frequency $f_G$ of the current level as an auxiliary guidance for context modeling. All the context information is then concatenated and fed into a tiny 2-layer MLP named \emph{Context Fuser} $\bm{C_p}$ to estimate the probability $p_i$ at vertex $n_i$. 

It is worth noting that if the number of previous layers is less than $L_c$, we set $L_c=l-1$ (\ie, using all the available previous layers for context). For level $l=1$, we only utilize its occurrence frequency $f_G^1$ to estimate the bit consumption because there is no previous layer. 

\begin{figure}
    \centering
    \includegraphics[width=0.9\linewidth]{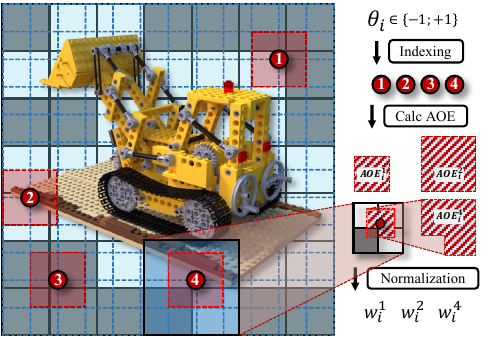}
    \caption{Illustration of Hash Fusion. 
    In this toy example, the resolutions of the voxel and occupancy grids are 12 and 7, respectively. The weight of each hash collided vertex $k$ of $\theta_i$ is normalized from its AOE, $AOE_i^k$, which measures the intersection area between the vertex grid \textcolor{hash_fusion_red}{(semitransparent dashed red square around the vertex)} and occupied cells (light-colored).
    }
    \vspace{-15pt}
    \label{fig:hash_fusion}
\end{figure}

\subsection{Hash Fusion with Occupancy Grid}
\label{subsec:hash_fusion}
One important design to alleviate extensive storage at the finer resolution of the trainable embeddings $\Theta$ is adopting spatial hashing indexing~\cite{INGP, teschner2003optimized}. However, this introduces an issue of hash collision when building the context models. Here, we provide a solution to address it for regression of more accurate predictions, 
with the assistance of the occupancy grid. 
The occupancy grid plays a pivotal role in our approach, which partitions the entire 3D scene into grid cells and records occupancy conditions in binary format. Generally, only cells on the surfaces of the objects are occupied, while the rest are empty, resulting in the sparsity. This inherent sparsity makes the occupancy grid a spatial prior that greatly enhances our context modeling.

Fig.~\ref{fig:hash_fusion} illustrates the details of the proposed hash fusion solution to address the hash collision issue. Particularly, suppose a feature vector $\theta_i$ corresponds to $K$ vertices, denoted as $\{n_i^k| k=1, \dots, K\}$ (\eg, $K=4$ in Fig.~\ref{fig:hash_fusion}). This implies for each $\theta_i$, multiple probabilities $\{p_i^k| k=1, \dots, K\}$ will be estimated. We define the Area of Effect (AOE) of a vertex as the intersection between the surrounded voxel grid and the occupied cell to weigh the probability prediction, 
which effectively determines the significance of a vertex. For example, vertex $n_i^3$ in Fig.~\ref{fig:hash_fusion} has an AOE of 0, then it is invalid, and should not contribute to the calculation of $p_i$. The final weighted probability of $p_i$ is expressed as
\vspace{-6pt}
\begin{equation}
\label{eq:hash_fusion}
    \begin{aligned}
    p_i &= \sum_{k=1}^{K}w_i^k \times p_i^k\\
    w_i^k &= AOE_i^k/\sum_{j=1}^{K}AOE_i^j
    \end{aligned}
\end{equation}
This not only enhances the training efficiency but also improves the context accuracy. Furthermore, if all associated vertices are invalid, then the corresponding $\theta_i$ is invalid and can directly be discarded to save storage space.

\subsection{Dimension-Wise Context Models}
\label{subsec:dimension_wise_context_models}
Considering BiRF introduces hybrid 2D-3D feature embeddings to improve the reconstruction quality, besides modeling contextual dependencies at various levels, we also emphasize the importance of cross-dimensional relations. The main idea of dimension-wise context models is to leverage the inherent relationship between tri-plane features and voxel features. Through extensive experiments, we found that 2D tri-plane feature embeddings cannot provide sufficient contextual information to predict the probability of 3D voxelized feature embeddings, likely due to missing one dimension. 
Thus, we turn to a more natural approach, \ie, estimating the probability of 2D plane embeddings from the 3D context.
%
Specifically, we employ a dimension projection design, as illustrated in the \textcolor{main_method_orange}{dashed orange box} of Fig.~\ref{fig:main_method} (bottom-left). We first reconstruct the entire 3D voxel using the spatial hashing function. Then, we project this 3D voxel along three different axes and record the frequency of $+1$s along each axis direction to obtain 2D Projected Voxel Features (PVF). Here, we leverage the prior knowledge of valid 3D space by the occupancy grid during projection. 
If the AOE of a vertex is 0, then it will be omitted from the calculation during the projection. The PVF will serve as one additional ``previous level" context to estimate the probability $p_i$ for each 2D $\theta_i$. Notably, PVFs can be obtained from three distinct 2D planes, \ie, the $xy$, $xz$, and $yz$ planes. In our work, we only utilize 3D feature embeddings that correspond to the largest resolution to generate PVFs, as they contain the most informative data.

\begin{figure*}
    \centering
    \includegraphics[width=0.97\linewidth]{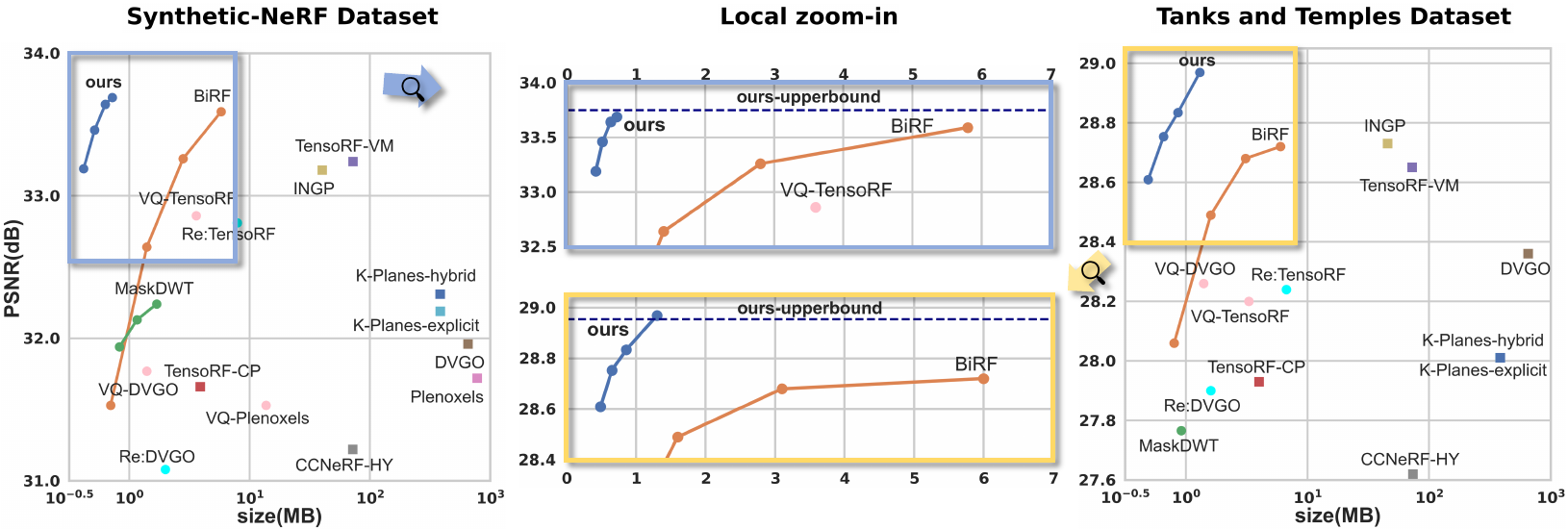}
    \caption{Performance overviews and detailed local zoom-in results of our proposed CNC and other methods. 
    We apply $\rm{log10}$ x-axis on the overviews for better visualization while linear x-axis on the zoom-in charts for better comparison. The more a curve goes upper-left, the better the rate-distortion (RD) performance is. Note that we achieve variable bitrates in our approach by changing $\lambda$ from $0.7e-3$ to $8e-3$, while BiRF~\cite{BiRF} achieves that by changing feature dimensions $F$ from 1 to 8. The dashed line ``ours-upperbound'' represents the upper fidelity bound of our binary NeRF model (\ie, $\lambda=0$). 
    }
    \vspace{-15pt}\label{fig:main_performance}
\end{figure*}

\noindent\textbf{Training loss}. With the establishment of our context models, we can calculate the entropy loss $L_{entropy}$, which is defined as the sum of the bits associated with all \emph{valid} feature vectors $\theta$s. The overall loss function then becomes
\vspace{-5pt}
\begin{equation}
\label{eq:loss}
    L = L_{mse} + \lambda L_{entropy} / M
\end{equation}
where $L_{mse}$ is the image reconstruction Mean Squared Error (MSE) loss and $\lambda$ is a tradeoff factor to balance the two terms for variable bitrates. $M$ is the number of $\theta$s in the embeddings, including both valid and invalid ones.

\noindent\textbf{Decoding and rendering process}. In the testing process, 3D embeddings are firstly decoded from shallow to deep levels using level-wise context models. Then the last 3D level is utilized to generate dimension-wise context for 2D embeddings. Finally, 2D embeddings are decoded in a coarse-to-fine order with the assistance of the dimension-wise context. It takes about 1 second for encoding/decoding. \emph{It's worth noting that once the embeddings are decoded, all the rendering processes are the same as INGP, requiring no additional time.}
\section{Experiments}
\label{sec:experiments}
In this section, we first describe the implementation details, then perform comparisons with previous methods on two benchmark datasets, and finally analyze different components of our CNC framework via extensive ablations. 
\subsection{Implementation Details}

Our model is implemented based on NerfAcc~\cite{NerfAcc} under PyTorch framework~\cite{pytorch} and is trained using a single NVIDIA RTX 3090 GPU. 
We use Adam optimizer~\cite{adam} with an initial learning rate of $0.01$ and train for $20000$ iterations.
For 3D embeddings, it contains 12 levels with resolutions from 16 to 512. 
For 2D embeddings, the resolutions range from 128 to 1024 with 4 levels. The numbers of maximum feature vectors are set to $2^{19}$ and $2^{17}$ per level for 3D and 2D, respectively. The resolution of the occupancy grid is 128. We set the feature vector dimension $F$ as 8, and the number of context levels $L_c$ as 3. The structure of the rendering MLP is the same as~\cite{INGP} but with a width of 160. During training, we vary $\lambda$ in Eq.~\ref{eq:hash_fusion} from $0.7e-3$ to $8e-3$ to obtain different bitrates. More details can be found in Sec.~\textcolor{red}{A} of the supplementary.

\subsection{Performance Evaluation}
\noindent\textbf{Baselines}. We mainly compared our method with the recent NeRF compression approaches.
Among them, BiRF~\cite{BiRF} and MaskDWT~\cite{MaskDWT} minimize NeRF model size during training, while VQRF~\cite{VQRF} and Re:NeRF~\cite{Re:NeRF} are post-training compression algorithms. We also compared several major variants of NeRF to see their storage cost, including DVGO~\cite{DVGO}, Plenoxels~\cite{Plenoxels}, TensoRF~\cite{TensoRF}, CCNeRF~\cite{CCNeRF}, INGP~\cite{INGP} and K-Planes~\cite{Kplanes}.


\noindent\textbf{Datasets}. Experiments are conducted on a synthetic dataset \emph{Synthetic-NeRF}~\cite{NeRF} and a real-world large-scale dataset \emph{Tanks and Temples}~\cite{Tanks_and_temples}. We follow the setting as BiRF~\cite{BiRF}.


\noindent\textbf{Metrics}. Besides the conventional PSNR versus size results, we also employ BD-rate~\cite{BD_rate} to assess the Rate-Distortion (RD) performance of these approaches, which measures the relative size change under the same fidelity quality. \emph{A reduced BD-rate signifies decreased bit consumption for the same quality}. 

\begin{figure*}
    \centering
    \includegraphics[width=0.95\linewidth]{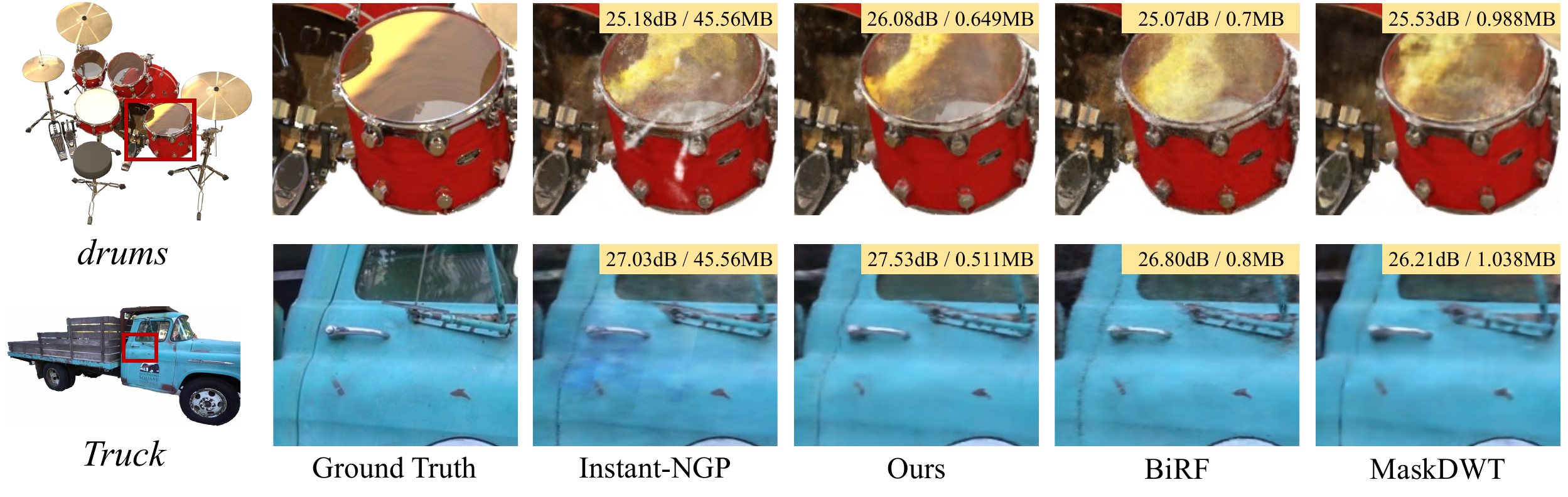}
    \caption{Qualitative quality comparisons of \textit{drums} in Synthetic-NeRF dataset and \textit{Truck} in Tanks and Temples dataset. We mainly compare recent NeRF compression approaches, along with our base model Instant-NGP. While some compression algorithms can achieve a low size of 1MB, they significantly sacrifice reconstruction fidelity. Our approach exhibits the best visual quality at the low size. Quantitative results of PSNR/size are shown in the upper right.}
    \vspace{-15pt}
    \label{fig:qualitative_comparison}
\end{figure*}

\noindent\textbf{Results}. We report the quantitative and qualitative results in Fig.~\ref{fig:main_performance} and Fig.~\ref{fig:qualitative_comparison}, respectively. For more fidelity metrics (SSIM~\cite{ssim} and LPIPS~\cite{LPIPS}) and visual comparisons, please refer to Tab.~\textcolor{red}{B-C} and Fig.~\textcolor{red}{A-B} of the the supplementary. Our proposed CNC achieves a significant RD performance advantage over others. \emph{Compared to the SOTA (\ie, BiRF), our proposed CNC achieves 86.7\% and 82.3\% BD-rate reduction on the two datasets.} For Synthetic-NeRF dataset, our CNC closely approaches the upper fidelity bound while maintaining a much smaller size, showcasing the effectiveness of CNC. For the Tanks and Temples dataset, our CNC even surpasses the upper-bound. We conjecture that, to some extent, the entropy constraint from the context models serves as regularization to prevent overfitting. 

\noindent\textbf{Bitstreams}. Our bitstream comprises four components: 3D and 2D feature embeddings, the rendering MLP, context models and the occupancy grid. Their average sizes are 0.220MB, 0.148MB, 0.011MB and 0.039MB in Synthetic-NeRF dataset with $\lambda=4e-3$. They are encoded/stored as follows. 
Feature embeddings are entropy encoded by Arithmetic Coding (AE)~\cite{AE} with probabilities predicted by context models. The rendering MLP parameters are quantized from the original 32 bits to 13 bits, which only causes a slight performance drop of less than 0.02 dB in PSNR while saving up to 0.216 MB.  
Context models are preserved in float32 to maintain prediction accuracy. The occupancy grid is binary and can be compressed by AE~\cite{AE}. 


\begin{table}
\setlength\tabcolsep{3pt}
    \centering \small
    \begin{tabular}{cccc}
    \toprule[2pt]
       \textbf{2D context} & \textbf{3D context} & \textbf{Dimension}  & \textbf{BD-rate(all/emb)} \\
    \midrule[1pt]
       \ding{51}  & \ding{51} & \ding{51} & 0\%/0\% \\
       \ding{51}  & \ding{51} & \ding{55} & +5.7\%/+9.2\% \\
       \ding{51}  & \ding{55} & \ding{55} & +30.8\%/+54.3\% \\
       \ding{55}  & \ding{55} & \ding{55} & +43.7\%/+78.8\% \\
    \bottomrule[2pt]
    \end{tabular}
    \caption{Ablation study on context models on Synthetic-NeRF dataset. Compared to CNC, disabling context models results in undesirable increases in BD-rate. ``BD-rate (all/emb)'' denotes the relative size changes in terms of the total model size or the size of the embeddings only. 
    }
    \label{tab:ablation_on_context_models}
\end{table}

\begin{figure}
    \centering
    \includegraphics[width=0.90\linewidth]{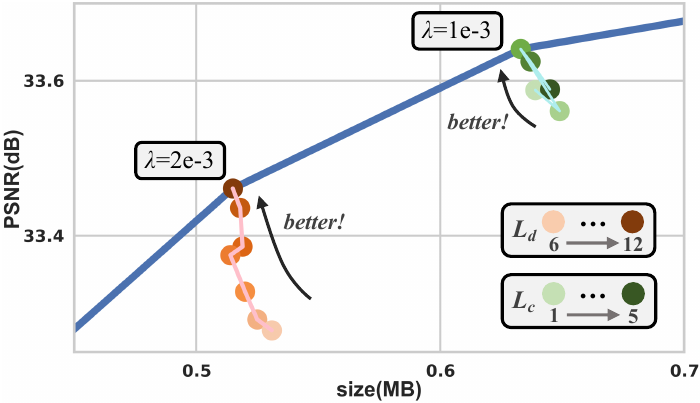}
    \caption{\textcolor{orange}{Orange points} represent ablation studies on $L_d$ from 6 to 12, with $\lambda=2e-3$. \textcolor{Ld_Lc_green}{Green points} represent ablation studies on $L_c$ from 1 to 5, with $\lambda=1e-3$. Best results are obtained at $L_d=12$ and $L_c=3$.
    Experiments are on Synthetic-NeRF dataset.}
    \vspace{-15pt}
    \label{fig:level_context}
\end{figure}

\begin{figure}
    \centering
    \includegraphics[width=0.93\linewidth]{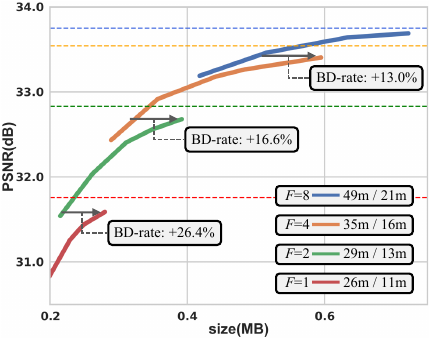}
    \caption{Fidelity upper-bound influences the RD performance. Dashed lines indicate the fidelity upperbounds at different hyperparameter settings of $F$. We also report the training time of our model with/without context models (bottom-right).
    }
\vspace{-0.2in}\label{fig:fidelity_upper_bound}
\end{figure}

\subsection{Ablation Study}


We contemplate what the optimal design is for context models. To address this question, we first deactivate certain context models to observe the extent of performance drop. Then, we delve into the detailed effect of inter-level dependencies in level-wise context models. Regarding the hash fusion module, we explore the crucial function of AOE, which can address the hash collision issue. 

\noindent\textbf{Which context model is the most useful?} First of all, we evaluate the capabilities of context models by disabling level-wise (3D and 2D embeddings) or dimension-wise context models. When context models are disabled, occurrence frequency $f_{G}^l$ is utilized to estimate all vector $\theta$s in embeddings for each level $l$. 
Note that $f_{G}^l$ is updated with the training. 
The corresponding results are shown in Tab.~\ref{tab:ablation_on_context_models}. It can be observed that a lack of either context model leads to a significant BD-rate increase. 3D context models contribute more than 2D ones since they occupy more storage space and are more sparse, thus having more potential for compression. Even though context models themselves introduce extra bits, 
the savings they bring in feature embeddings are remarkable, 
thanks to the accurate prediction of the probabilities.

\noindent\textbf{How to design the level-wise context model?} We now delve into the contribution of context models from each level. 
Specifically, we gradually replace the context model with $f_{G}^l$ from deeper to shallower layers, where we use $L_d$ to indicate the level starting from which context models are disabled. 
Experimental results are shown in Fig.~\ref{fig:level_context} by \textcolor{orange}{orange points}. We observe that as $L_d$ becomes smaller, RD performance decreases. This suggests that a single $f_{G}^l$ is inadequate to predict the feature distribution for each level $l$. In contrast, our context models exhibit greater capability in context aggregation. 
Experiments on context levels $L_c$ are also conducted, as shown in \textcolor{Ld_Lc_green}{green points} in Fig.~\ref{fig:level_context}. Increasing $L_c$ does not always lead to improved performance, as a distant level may provide limited information but introduce additional complexity.

\noindent\textbf{Which contextual order is suitable?} We investigate the order of fine-to-coarse in level-wise context models in Tab.~\ref{tab:effect_of_hash_fusion}. 
It can be seen that the coarse-to-fine context models performs much better than the reverse one. 
This suggests a coarse-to-fine flow aligns better with the information restoration behavior for a multi-resolution structure.

\begin{table}
    \setlength\tabcolsep{3pt}
    \centering \small
    \begin{tabular}{lcc}
    \toprule[2pt]
       \textbf{Ablation items} & \textbf{BD-rate(all/emb)} \\
    \midrule[1pt]
       Fine-to-coarse level-wise contexts & +33.8\%/+53.0\% \\
    \midrule[1pt]
       No discarding of $\theta$s & +43.7\%/+78.8\% \\
       Proper discarding of $\theta$s & +32.6\%/+59.1\% \\
       Over discarding of $\theta$s & N/A \\
    \bottomrule[2pt]
    \end{tabular}
    \caption{Ablation study on context dependencies and hash fusion on Synthetic-NeRF dataset.} 
    \vspace{-15pt}
    \label{tab:effect_of_hash_fusion}
\end{table}

\noindent\textbf{To which extent should invalid vectors be discarded in hash fusion?} Lastly, we conduct experiments to assess the effectiveness of hash fusion, for which a key function is to discard invalid feature vectors using AOEs to save storage space. To demonstrate its effectiveness, we vary the extent of discarding to observe the impact on RD performance. For ablation purposes, we disable both level-wise and dimension-wise context models and only use the frequency $f_G^l$ to estimate probabilities for each level $l$. The results are presented in Tab.~\ref{tab:effect_of_hash_fusion}. Initially, \emph{no discarding of $\theta$s}: we do not discard any of the feature vector $\theta$s and retain all of them, leading to significant storage waste on invalid vectors. This setting is the same as the last line of Tab.~\ref{tab:ablation_on_context_models}. Moving one step further, \emph{proper discarding of $\theta$s}: we apply $f_{G}^l$s only to valid $\theta$s at each level $l$ and encode them, whose validity is judged by AOEs. This approach aligns with our current methodology. Finally, \emph{over discarding of $\theta$s}: we alter the criterion by simply determining the validity of $\theta$ based on whether it is located in an occupied cell, rather than using AOE. However, this may cause over-discarding, where vertices necessary for rendering might be undecodable. This leads to a significant degradation in fidelity to an extremely low level (approximately 27.2 dB in PSNR under $\lambda=0.7e-3$), resulting in no intersection on the y-axis, making BD-rate incalculable (N/A). 


\subsection{Fidelity Upper-Bound Influences Performance}
In this subsection, we delve into a fundamental difference between NeRF compression and other data formats (such as image compression). To be specific, the ground-truth image for image compression is always available, which theoretically allows for perfect fidelity if no entropy constraint is applied. However, this is not the case for NeRF compression.
The ground truth 3D scene is not known in advance, and the upper-bound of the fidelity is fundamentally determined by the capability of the reconstruction algorithm. In our case, it is the CNC model without the entropy constraint, \ie, $\lambda=0$.  
 Fig.~\ref{fig:fidelity_upper_bound} shows our fidelity upperbounds under different feature dimensions $F$, ranging from 1 to 8.  
We can see that larger feature dimensions result in higher fidelity upperbounds and better RD performance.
This is because a larger feature dimension allows more room for context models to eliminate redundancy and perform compression. 
However, a higher upper-bound also leads to increased training and rendering time, and compression becomes more challenging when approaching the upper-bound.

\section{Conclusion}
\label{sec:conclusion}

In this paper, we have proposed a Context-based NeRF Compression (CNC) framework, where context models are carefully designed to eliminate the redundancy of binarized embeddings. Hash collision and occupancy grid are also fully exploited to further improve prediction accuracy. Experimental results on two benchmark datasets have demonstrated that our CNC can significantly compress multi-resolution Instant-NGP-based NeRFs and achieve SOTA performance. The success of NeRF compression on static scenes provides a solid proof of concept for more advanced and space-taking applications such as dynamic or large-scale NeRFs.

\noindent\textbf{Limitation}. The main drawback of our approach is the slowdown in training time, resulting in about 1.3$\times$ longer training duration over the one without context models. However, this limitation can be mitigated by: 1) reducing fidelity upper-bound; 2) adjusting context models; 
3) improving the code to execute context models and the rendering MLP concurrently. 


\section*{Acknowledgement}
\noindent The paper is supported in part by The National Natural Science Foundation of China (No. U21B2013).

\noindent MH is supported by funding from The Australian Research Council Discovery Program DP230101176.

{
    \small
    \bibliographystyle{ieeenat_fullname}
    \bibliography{main}

\begin{thebibliography}{52}
\providecommand{\natexlab}[1]{#1}
\providecommand{\url}[1]{\texttt{#1}}
\expandafter\ifx\csname urlstyle\endcsname\relax
  \providecommand{\doi}[1]{doi: #1}\else
  \providecommand{\doi}{doi: \begingroup \urlstyle{rm}\Url}\fi

\bibitem[Barron et~al.(2023)Barron, Mildenhall, Verbin, Srinivasan, and Hedman]{barron2023zip}
Jonathan~T. Barron, Ben Mildenhall, Dor Verbin, Pratul~P. Srinivasan, and Peter Hedman.
\newblock Zip-nerf: Anti-aliased grid-based neural radiance fields.
\newblock \emph{ICCV}, 2023.

\bibitem[Bengio et~al.(2013)Bengio, L{\'e}onard, and Courville]{ste}
Yoshua Bengio, Nicholas L{\'e}onard, and Aaron Courville.
\newblock Estimating or propagating gradients through stochastic neurons for conditional computation.
\newblock \emph{arXiv preprint arXiv:1308.3432}, 2013.

\bibitem[Bjontegaard(2001)]{BD_rate}
Gisle Bjontegaard.
\newblock Calculation of average psnr differences between rd-curves.
\newblock \emph{ITU SG16 Doc. VCEG-M33}, 2001.

\bibitem[Chen et~al.(2022)Chen, Xu, Geiger, Yu, and Su]{TensoRF}
Anpei Chen, Zexiang Xu, Andreas Geiger, Jingyi Yu, and Hao Su.
\newblock Tensorf: Tensorial radiance fields.
\newblock In \emph{European Conference on Computer Vision}, pages 333--350. Springer, 2022.

\bibitem[Cheng et~al.(2020)Cheng, Sun, Takeuchi, and Katto]{cheng2020learned}
Zhengxue Cheng, Heming Sun, Masaru Takeuchi, and Jiro Katto.
\newblock Learned image compression with discretized gaussian mixture likelihoods and attention modules.
\newblock In \emph{Proceedings of the IEEE/CVF conference on computer vision and pattern recognition}, pages 7939--7948, 2020.

\bibitem[Deng and Tartaglione(2023)]{Re:NeRF}
Chenxi~Lola Deng and Enzo Tartaglione.
\newblock Compressing explicit voxel grid representations: fast nerfs become also small.
\newblock In \emph{Proceedings of the IEEE/CVF Winter Conference on Applications of Computer Vision}, pages 1236--1245, 2023.

\bibitem[Fridovich-Keil et~al.(2022)Fridovich-Keil, Yu, Tancik, Chen, Recht, and Kanazawa]{Plenoxels}
Sara Fridovich-Keil, Alex Yu, Matthew Tancik, Qinhong Chen, Benjamin Recht, and Angjoo Kanazawa.
\newblock Plenoxels: Radiance fields without neural networks.
\newblock In \emph{Proceedings of the IEEE/CVF Conference on Computer Vision and Pattern Recognition}, pages 5501--5510, 2022.

\bibitem[Fridovich-Keil et~al.(2023)Fridovich-Keil, Meanti, Warburg, Recht, and Kanazawa]{Kplanes}
Sara Fridovich-Keil, Giacomo Meanti, Frederik~Rahb{\ae}k Warburg, Benjamin Recht, and Angjoo Kanazawa.
\newblock K-planes: Explicit radiance fields in space, time, and appearance.
\newblock In \emph{Proceedings of the IEEE/CVF Conference on Computer Vision and Pattern Recognition}, pages 12479--12488, 2023.

\bibitem[Girish et~al.(2023)Girish, Shrivastava, and Gupta]{SHACIRA}
Sharath Girish, Abhinav Shrivastava, and Kamal Gupta.
\newblock Shacira: Scalable hash-grid compression for implicit neural representations.
\newblock In \emph{Proceedings of the IEEE/CVF International Conference on Computer Vision}, pages 17513--17524, 2023.

\bibitem[Gou et~al.(2021)Gou, Yu, Maybank, and Tao]{distillation}
Jianping Gou, Baosheng Yu, Stephen~J Maybank, and Dacheng Tao.
\newblock Knowledge distillation: A survey.
\newblock \emph{International Journal of Computer Vision}, 129:\penalty0 1789--1819, 2021.

\bibitem[Han et~al.(2016)Han, Mao, and Dally]{deepcompression}
Song Han, Huizi Mao, and William~J Dally.
\newblock Deep compression: Compressing deep neural networks with pruning, trained quantization and huffman coding.
\newblock \emph{International Conference on Learning Representations (ICLR)}, 2016.

\bibitem[He et~al.(2021)He, Zheng, Sun, Wang, and Qin]{he2021checkerboard}
Dailan He, Yaoyan Zheng, Baocheng Sun, Yan Wang, and Hongwei Qin.
\newblock Checkerboard context model for efficient learned image compression.
\newblock In \emph{Proceedings of the IEEE/CVF Conference on Computer Vision and Pattern Recognition}, pages 14771--14780, 2021.

\bibitem[He et~al.(2022)He, Yang, Peng, Ma, Qin, and Wang]{he2022elic}
Dailan He, Ziming Yang, Weikun Peng, Rui Ma, Hongwei Qin, and Yan Wang.
\newblock Elic: Efficient learned image compression with unevenly grouped space-channel contextual adaptive coding.
\newblock In \emph{Proceedings of the IEEE/CVF Conference on Computer Vision and Pattern Recognition}, pages 5718--5727, 2022.

\bibitem[Jaderberg et~al.(2014)Jaderberg, Vedaldi, and Zisserman]{low-rank-1}
Max Jaderberg, Andrea Vedaldi, and Andrew Zisserman.
\newblock Speeding up convolutional neural networks with low rank expansions.
\newblock In \emph{Proceedings of the British Machine Vision Conference 2014}. British Machine Vision Association, 2014.

\bibitem[Kingma and Ba(2015)]{adam}
Diederik Kingma and Jimmy Ba.
\newblock Adam: A method for stochastic optimization.
\newblock In \emph{International Conference on Learning Representations (ICLR)}, San Diega, CA, USA, 2015.

\bibitem[Knapitsch et~al.(2017)Knapitsch, Park, Zhou, and Koltun]{Tanks_and_temples}
Arno Knapitsch, Jaesik Park, Qian-Yi Zhou, and Vladlen Koltun.
\newblock Tanks and temples: Benchmarking large-scale scene reconstruction.
\newblock \emph{ACM Transactions on Graphics (ToG)}, 36\penalty0 (4):\penalty0 1--13, 2017.

\bibitem[Kutulakos and Seitz(2000)]{kutulakos2000theory}
Kiriakos~N Kutulakos and Steven~M Seitz.
\newblock A theory of shape by space carving.
\newblock \emph{International journal of computer vision}, 38:\penalty0 199--218, 2000.

\bibitem[Li et~al.(2021)Li, Li, and Lu]{DCVC}
Jiahao Li, Bin Li, and Yan Lu.
\newblock Deep contextual video compression.
\newblock \emph{Advances in Neural Information Processing Systems}, 34:\penalty0 18114--18125, 2021.

\bibitem[Li et~al.(2022{\natexlab{a}})Li, Li, and Lu]{DCVC-HEM}
Jiahao Li, Bin Li, and Yan Lu.
\newblock Hybrid spatial-temporal entropy modelling for neural video compression.
\newblock In \emph{Proceedings of the 30th ACM International Conference on Multimedia}, pages 1503--1511, 2022{\natexlab{a}}.

\bibitem[Li et~al.(2023{\natexlab{a}})Li, Shen, Wang, Shen, and Bo]{VQRF}
Lingzhi Li, Zhen Shen, Zhongshu Wang, Li Shen, and Liefeng Bo.
\newblock Compressing volumetric radiance fields to 1 mb.
\newblock In \emph{Proceedings of the IEEE/CVF Conference on Computer Vision and Pattern Recognition}, pages 4222--4231, 2023{\natexlab{a}}.

\bibitem[Li et~al.(2023{\natexlab{b}})Li, Wang, Shen, Shen, and Tan]{CompactNeRF}
Lingzhi Li, Zhongshu Wang, Zhen Shen, Li Shen, and Ping Tan.
\newblock Compact real-time radiance fields with neural codebook.
\newblock In \emph{ICME}, 2023{\natexlab{b}}.

\bibitem[Li et~al.(2022{\natexlab{b}})Li, Tancik, and Kanazawa]{NerfAcc}
Ruilong Li, Matthew Tancik, and Angjoo Kanazawa.
\newblock Nerfacc: A general nerf acceleration toolbox.
\newblock \emph{arXiv preprint arXiv:2210.04847}, 2022{\natexlab{b}}.

\bibitem[Li et~al.(2023{\natexlab{c}})Li, Jiang, Xu, Xiangli, Wang, Lin, and Dai]{li2023matrixcity}
Yixuan Li, Lihan Jiang, Linning Xu, Yuanbo Xiangli, Zhenzhi Wang, Dahua Lin, and Bo Dai.
\newblock Matrixcity: A large-scale city dataset for city-scale neural rendering and beyond.
\newblock In \emph{Proceedings of the IEEE/CVF International Conference on Computer Vision}, pages 3205--3215, 2023{\natexlab{c}}.

\bibitem[Liu et~al.(2020)Liu, Gu, Zaw~Lin, Chua, and Theobalt]{liu2020neural}
Lingjie Liu, Jiatao Gu, Kyaw Zaw~Lin, Tat-Seng Chua, and Christian Theobalt.
\newblock Neural sparse voxel fields.
\newblock \emph{Advances in Neural Information Processing Systems}, 33:\penalty0 15651--15663, 2020.

\bibitem[Lombardi et~al.(2019)Lombardi, Simon, Saragih, Schwartz, Lehrmann, and Sheikh]{neuralvolume}
Stephen Lombardi, Tomas Simon, Jason Saragih, Gabriel Schwartz, Andreas Lehrmann, and Yaser Sheikh.
\newblock Neural volumes: Learning dynamic renderable volumes from images.
\newblock \emph{ACM Trans. Graph.}, 38\penalty0 (4):\penalty0 65:1--65:14, 2019.

\bibitem[Martel et~al.(2021)Martel, Lindell, Lin, Chan, Monteiro, and Wetzstein]{martel2021acorn}
Julien~NP Martel, David~B Lindell, Connor~Z Lin, Eric~R Chan, Marco Monteiro, and Gordon Wetzstein.
\newblock Acorn: adaptive coordinate networks for neural scene representation.
\newblock \emph{ACM Transactions on Graphics (TOG)}, 40\penalty0 (4):\penalty0 1--13, 2021.

\bibitem[Max(1995)]{volume_rendering}
Nelson Max.
\newblock Optical models for direct volume rendering.
\newblock \emph{IEEE Transactions on Visualization and Computer Graphics}, 1\penalty0 (2):\penalty0 99--108, 1995.

\bibitem[Mildenhall et~al.(2021)Mildenhall, Srinivasan, Tancik, Barron, Ramamoorthi, and Ng]{NeRF}
Ben Mildenhall, Pratul~P Srinivasan, Matthew Tancik, Jonathan~T Barron, Ravi Ramamoorthi, and Ren Ng.
\newblock Nerf: Representing scenes as neural radiance fields for view synthesis.
\newblock \emph{Communications of the ACM}, 65\penalty0 (1):\penalty0 99--106, 2021.

\bibitem[M{\"u}ller et~al.(2022)M{\"u}ller, Evans, Schied, and Keller]{INGP}
Thomas M{\"u}ller, Alex Evans, Christoph Schied, and Alexander Keller.
\newblock Instant neural graphics primitives with a multiresolution hash encoding.
\newblock \emph{ACM Transactions on Graphics (ToG)}, 41\penalty0 (4):\penalty0 1--15, 2022.

\bibitem[Paszke et~al.(2019)Paszke, Gross, Massa, Lerer, Bradbury, Chanan, Killeen, Lin, Gimelshein, Antiga, et~al.]{pytorch}
Adam Paszke, Sam Gross, Francisco Massa, Adam Lerer, James Bradbury, Gregory Chanan, Trevor Killeen, Zeming Lin, Natalia Gimelshein, Luca Antiga, et~al.
\newblock Pytorch: An imperative style, high-performance deep learning library.
\newblock \emph{Advances in neural information processing systems}, 32, 2019.

\bibitem[Polino et~al.(2018)Polino, Pascanu, and Alistarh]{model_quantization}
Antonio Polino, Razvan Pascanu, and Dan Alistarh.
\newblock Model compression via distillation and quantization.
\newblock In \emph{International Conference on Learning Representations}, 2018.

\bibitem[Rho et~al.(2023)Rho, Lee, Nam, Lee, Ko, and Park]{MaskDWT}
Daniel Rho, Byeonghyeon Lee, Seungtae Nam, Joo~Chan Lee, Jong~Hwan Ko, and Eunbyung Park.
\newblock Masked wavelet representation for compact neural radiance fields.
\newblock In \emph{Proceedings of the IEEE/CVF Conference on Computer Vision and Pattern Recognition}, pages 20680--20690, 2023.

\bibitem[Rigamonti et~al.(2013)Rigamonti, Sironi, Lepetit, and Fua]{low-rank-2}
Roberto Rigamonti, Amos Sironi, Vincent Lepetit, and Pascal Fua.
\newblock Learning separable filters.
\newblock In \emph{Proceedings of the IEEE conference on computer vision and pattern recognition}, pages 2754--2761, 2013.

\bibitem[Seitz and Dyer(1999)]{seitz1999photorealistic}
Steven~M Seitz and Charles~R Dyer.
\newblock Photorealistic scene reconstruction by voxel coloring.
\newblock \emph{International Journal of Computer Vision}, 35:\penalty0 151--173, 1999.

\bibitem[Sheng et~al.(2022)Sheng, Li, Li, Li, Liu, and Lu]{DCVC-TCM}
Xihua Sheng, Jiahao Li, Bin Li, Li Li, Dong Liu, and Yan Lu.
\newblock Temporal context mining for learned video compression.
\newblock \emph{IEEE Transactions on Multimedia}, 2022.

\bibitem[Shin and Park(2023)]{BiRF}
Seungjoo Shin and Jaesik Park.
\newblock Binary radiance fields.
\newblock \emph{Advances in neural information processing systems}, 2023.

\bibitem[Sitzmann et~al.(2019)Sitzmann, Thies, Heide, Nie{\ss}ner, Wetzstein, and Zollhofer]{sitzmann2019deepvoxels}
Vincent Sitzmann, Justus Thies, Felix Heide, Matthias Nie{\ss}ner, Gordon Wetzstein, and Michael Zollhofer.
\newblock Deepvoxels: Learning persistent 3d feature embeddings.
\newblock In \emph{Proceedings of the IEEE/CVF Conference on Computer Vision and Pattern Recognition}, pages 2437--2446, 2019.

\bibitem[Sun et~al.(2022)Sun, Sun, and Chen]{DVGO}
Cheng Sun, Min Sun, and Hwann-Tzong Chen.
\newblock Direct voxel grid optimization: Super-fast convergence for radiance fields reconstruction.
\newblock In \emph{Proceedings of the IEEE/CVF Conference on Computer Vision and Pattern Recognition}, pages 5459--5469, 2022.

\bibitem[Takikawa et~al.(2021)Takikawa, Litalien, Yin, Kreis, Loop, Nowrouzezahrai, Jacobson, McGuire, and Fidler]{takikawa2021neural}
Towaki Takikawa, Joey Litalien, Kangxue Yin, Karsten Kreis, Charles Loop, Derek Nowrouzezahrai, Alec Jacobson, Morgan McGuire, and Sanja Fidler.
\newblock Neural geometric level of detail: Real-time rendering with implicit 3d shapes.
\newblock In \emph{Proceedings of the IEEE/CVF Conference on Computer Vision and Pattern Recognition}, pages 11358--11367, 2021.

\bibitem[Takikawa et~al.(2022)Takikawa, Evans, Tremblay, M{\"u}ller, McGuire, Jacobson, and Fidler]{takikawa2022variable}
Towaki Takikawa, Alex Evans, Jonathan Tremblay, Thomas M{\"u}ller, Morgan McGuire, Alec Jacobson, and Sanja Fidler.
\newblock Variable bitrate neural fields.
\newblock In \emph{ACM SIGGRAPH 2022 Conference Proceedings}, pages 1--9, 2022.

\bibitem[Tancik et~al.(2020)Tancik, Srinivasan, Mildenhall, Fridovich-Keil, Raghavan, Singhal, Ramamoorthi, Barron, and Ng]{tancik2020fourier}
Matthew Tancik, Pratul Srinivasan, Ben Mildenhall, Sara Fridovich-Keil, Nithin Raghavan, Utkarsh Singhal, Ravi Ramamoorthi, Jonathan Barron, and Ren Ng.
\newblock Fourier features let networks learn high frequency functions in low dimensional domains.
\newblock \emph{Advances in Neural Information Processing Systems}, 33:\penalty0 7537--7547, 2020.

\bibitem[Tancik et~al.(2022)Tancik, Casser, Yan, Pradhan, Mildenhall, Srinivasan, Barron, and Kretzschmar]{tancik2022block}
Matthew Tancik, Vincent Casser, Xinchen Yan, Sabeek Pradhan, Ben Mildenhall, Pratul~P Srinivasan, Jonathan~T Barron, and Henrik Kretzschmar.
\newblock Block-nerf: Scalable large scene neural view synthesis.
\newblock In \emph{Proceedings of the IEEE/CVF Conference on Computer Vision and Pattern Recognition}, pages 8248--8258, 2022.

\bibitem[Tancik et~al.(2023)Tancik, Weber, Ng, Li, Yi, Wang, Kristoffersen, Austin, Salahi, Ahuja, et~al.]{tancik2023nerfstudio}
Matthew Tancik, Ethan Weber, Evonne Ng, Ruilong Li, Brent Yi, Terrance Wang, Alexander Kristoffersen, Jake Austin, Kamyar Salahi, Abhik Ahuja, et~al.
\newblock Nerfstudio: A modular framework for neural radiance field development.
\newblock In \emph{ACM SIGGRAPH 2023 Conference Proceedings}, pages 1--12, 2023.

\bibitem[Tang et~al.(2022)Tang, Chen, Wang, and Zeng]{CCNeRF}
Jiaxiang Tang, Xiaokang Chen, Jingbo Wang, and Gang Zeng.
\newblock Compressible-composable nerf via rank-residual decomposition.
\newblock \emph{Advances in Neural Information Processing Systems}, 35:\penalty0 14798--14809, 2022.

\bibitem[Teschner et~al.(2003)Teschner, Heidelberger, M{\"u}ller, Pomerantes, and Gross]{teschner2003optimized}
Matthias Teschner, Bruno Heidelberger, Matthias M{\"u}ller, Danat Pomerantes, and Markus~H Gross.
\newblock Optimized spatial hashing for collision detection of deformable objects.
\newblock In \emph{Vmv}, pages 47--54, 2003.

\bibitem[Wang et~al.(2004)Wang, Bovik, Sheikh, and Simoncelli]{ssim}
Zhou Wang, Alan~C Bovik, Hamid~R Sheikh, and Eero~P Simoncelli.
\newblock Image quality assessment: from error visibility to structural similarity.
\newblock \emph{IEEE transactions on image processing}, 13\penalty0 (4):\penalty0 600--612, 2004.

\bibitem[Witten et~al.(1987)Witten, Neal, and Cleary]{AE}
Ian~H Witten, Radford~M Neal, and John~G Cleary.
\newblock Arithmetic coding for data compression.
\newblock \emph{Communications of the ACM}, 30\penalty0 (6):\penalty0 520--540, 1987.

\bibitem[Xiangli et~al.(2022)Xiangli, Xu, Pan, Zhao, Rao, Theobalt, Dai, and Lin]{CityNeRF}
Yuanbo Xiangli, Linning Xu, Xingang Pan, Nanxuan Zhao, Anyi Rao, Christian Theobalt, Bo Dai, and Dahua Lin.
\newblock Bungeenerf: Progressive neural radiance field for extreme multi-scale scene rendering.
\newblock In \emph{European conference on computer vision}, pages 106--122. Springer, 2022.

\bibitem[Yang et~al.(2023)Yang, Pavone, and Wang]{yang2023freenerf}
Jiawei Yang, Marco Pavone, and Yue Wang.
\newblock Freenerf: Improving few-shot neural rendering with free frequency regularization.
\newblock In \emph{Proceedings of the IEEE/CVF Conference on Computer Vision and Pattern Recognition}, pages 8254--8263, 2023.

\bibitem[Yu et~al.(2021)Yu, Li, Tancik, Li, Ng, and Kanazawa]{yu2021plenoctrees}
Alex Yu, Ruilong Li, Matthew Tancik, Hao Li, Ren Ng, and Angjoo Kanazawa.
\newblock Plenoctrees for real-time rendering of neural radiance fields.
\newblock In \emph{Proceedings of the IEEE/CVF International Conference on Computer Vision}, pages 5752--5761, 2021.

\bibitem[Zhang et~al.(2018)Zhang, Isola, Efros, Shechtman, and Wang]{LPIPS}
Richard Zhang, Phillip Isola, Alexei~A Efros, Eli Shechtman, and Oliver Wang.
\newblock The unreasonable effectiveness of deep features as a perceptual metric.
\newblock In \emph{Proceedings of the IEEE conference on computer vision and pattern recognition}, pages 586--595, 2018.

\bibitem[Zhu and Gupta(2018)]{zhu2017prune}
Michael Zhu and Suyog Gupta.
\newblock To prune, or not to prune: exploring the efficacy of pruning for model compression.
\newblock In \emph{International Conference on Learning Representations (ICLR)}, Vancouver, CANADA, 2018.

\end{thebibliography}
}

\clearpage
\setcounter{page}{1}
\maketitlesupplementary

\renewcommand\thesection{\Alph{section}}
\renewcommand\thetable{\Alph{table}}
\renewcommand\thefigure{\Alph{figure}}
\setcounter{section}{0}
\setcounter{table}{0}
\setcounter{figure}{0}

\noindent In this supplementary material, we first report more implementation details in Sec.~\ref{sec:more_implementation_details}, then exhibit the efficient backward of our bit estimator function in Sec.~\ref{sec:efficient_bit_estimator_backward}, and we also present a notation table in Sec.~\ref{sec:notation_table} for clarity in understanding our paper. Additionally, more quantitative and qualitative results are included at the end of the document.

\section{More Implementation Details}
\label{sec:more_implementation_details}


\noindent\textbf{Context Fuser}. \textit{Context Fuser} is able to aggregate contexts from previous $L_c$ levels. For 3D embeddings, it is a 3-layer MLP. It has an input channel of $F\times L_c+1$, a hidden channel of $32$ and an output channel of $F$ with $\rm{LeakyReLU}$ activation, where ``$+1$'' is for $f_G$. For 2D embeddings, it is a 1-layer linear module. It has an input channel of $F\times L_c+1+F$ and an output channel of $F$, where ``$+F$'' is for dimension-wise context. Note that different levels of the same $L_c$s share one \textit{Context Fuser} to save storage space.

\noindent\textbf{Learning Rate}. Our initial learning rate is 0.01 and the total iteration is $20K$. For the first $1K$ iterations, we adopt a linear warm-up stage to stabilize the training process. For the rest iterations, the learning rate is decayed by a factor of 0.33 at $9K, 12K, 15K, 17K$ and $19K$ iterations.

\noindent\textbf{Sampling Strategy}. During training, feeding all embeddings to the context models in a single iteration will lead to out-of-memory (OOM). To address this, we randomly sample $150K$ feature vectors $\theta$s for training 3D embeddings under $F=8$ in each iteration, and $200K$ under $F=1,2,4$. For 2D embeddings, we do not employ this strategy but feed them all together in one iteration.

\noindent\textbf{Quantization of the Rendering MLP}. We utilize 13 bits to quantize the rendering MLP. For each parameter $\omega_i\in \Omega$,
\vspace{-7pt}
\begin{equation}
    \omega_{qi} = \lfloor(\omega_i-\rm{MIN}(\Omega))\frac{2^D-1}{\rm{MAX}(\Omega)-\rm{MIN}(\Omega)}\rfloor
\end{equation}
where $\Omega$ is the parameter collection of the rendering MLP and $\omega_{qi}$ is the quantized parameter. $D=13$ represents the number of digits for quantization. $\rm{MIN}$ and $\rm{MAX}$ represent operations to calculate minimum and maximum elements, respectively.

\noindent\textbf{Inverse Hash Mapping}. While the hash function~\cite{INGP} provides only a unidirectional mapping of $n\to\theta$, we are in need of its inverse mapping of $\theta\to n$.
To accomplish this, during the initialization stage, we traverse all $n$s in voxels using the hash function and store their corresponding $\theta$s, which takes a GPU memory of 5 GB. Consequently, we can retrieve all associated vertices $\{n_i^k | k=1, \dots, K\}$ of a random vector $\theta_i$ by querying this recorded information during training.

\section{Efficient Backward of Bit Estimator}
\label{sec:efficient_bit_estimator_backward}

In this paper, we estimate the bit consumption of a $\theta_i$ with its probability $p_i$ in a differentiable formula, as shown below (same as Eq.~\ref{eq:inforamtion_entropy}):
\begin{equation}
\label{eq:inforamtion_entropy_supple}
    \textit{bit}(p_i|\theta_i)
    = -(\frac{1+\theta_i}{2}\log_2(p_i)+\frac{1-\theta_i}{2}\log_2(1-p_i))
\end{equation}

\noindent We discover this estimator is better than the below one:
\begin{equation}
\label{eq:inforamtion_entropy_old_supple}
    \textit{bit}(p_i|\theta_i)
    = -\log_2(\frac{1-\theta_i}{2}+\theta_ip_i)
\end{equation}
Although these two forms of estimator functions produce the same results of bit consumption in their forward pass, they exhibit quite different behavior for backpropagation. For Eq.~\ref{eq:inforamtion_entropy_supple},

\begin{equation}
\left\{
\begin{aligned}
    \frac{\partial bit}{\partial \theta_i} =& 
    \frac{1}{2}\log_2(\frac{1}{p_i}-1) \\
    \frac{\partial bit}{\partial p_i} =& 
    \left\{
        \begin{array}{ll}
            -\frac{1}{p_i\ln2} & \quad \theta_i = +1 \\
            -\frac{1}{(p_i-1)\ln2} & \quad \theta_i = -1
        \end{array}
    \right.
\end{aligned}
\right.
\end{equation}

\noindent However, there exists a different derivative formula of $\theta_i$ for Eq.~\ref{eq:inforamtion_entropy_old_supple},
\begin{equation}
\left\{
\begin{aligned}
    \frac{\partial bit}{\partial \theta_i} =& 
    \left\{
        \begin{array}{ll}
            -\frac{p_i-0.5}{p_i\ln2} & \quad \theta_i = +1 \\
            -\frac{p_i-0.5}{(1-p_i)\ln2} & \quad \theta_i = -1
        \end{array}
    \right. \\
    \frac{\partial bit}{\partial p_i} =& 
    \left\{
        \begin{array}{ll}
            -\frac{1}{p_i\ln2} & \quad \theta_i = +1 \\
            -\frac{1}{(p_i-1)\ln2} & \quad \theta_i = -1
        \end{array}
    \right.
\end{aligned}
\right.
\end{equation}

\noindent This inherent difference of backward propagation of Eq.~\ref{eq:inforamtion_entropy_old_supple} results in a more intricate gradient, significantly slowing down the training speed to 68 minutes. Additionally, it exacerbates the RD performance, leading to undesirable BD-rate increases of +49.3\%/+89.5\%.

\section{Notation Table and More Results}
\label{sec:notation_table}
In this subsection, we first provide a notation table, which is necessary for readers to understand our paper. Subsequently, we showcase additional quantitative and qualitative results, offering more detailed data for thorough understanding.


\begin{table*} 
    \centering
    \begin{tabular}{cl}
    \toprule[2pt]
      \textbf{Notation}   & \textbf{Definition} \\
    \midrule[1pt]
        $\mathbf{x}$ &An input coordinate for rendering \\ 
        $\mathbf{d}$ &Viewing direction of the input coordinate $\mathbf{x}$\\ 
        $\mathbf{o}$ &Camera center to observe the input coordinate $\mathbf{x}$\\ 
        $\mathbf{r}$ &A ray for rendering\\ 
        $v$ &Index of a sampled point along the ray $\mathbf{r}$\\ 
        $\sigma$ &Density of the sampled point $v$\\
        $\mathbf{c}$ &Color of the sampled point $v$\\ 
        $T$ &Transmittance to the sampled point $v$ along the ray $\mathbf{r}$\\
        $\hat{C}$ &The rendered pixel color of the ray $\mathbf{r}$\\
        $\mathbf{f}$ &The interpolated input feature for positional encoding\\
        \cline{1-2}
        $L$ &Total resolution level number of embeddings \\
        $l$ &A level out of $L$ \\
        $\Theta$ &Collection of feature embeddings in one level \\
        $\bm{\theta}$ &A vector element of embeddings $\Theta$ \\
        $\theta$ &A scalar of $\bm{\theta}$, which can be either $-1$ or $+1$ \\
        $i$ &Index of a randomly sampled $\theta$ \\
        $T$ &Size of embeddings $\Theta$ \\
        $f_G$ &Occurrence frequency of $+1$ in embeddings $\Theta$ \\
        $n$ &Associated vertex of $\theta$ in the voxel \\
        $p$ &Estimated probability for entropy modeling \\
        $L_c$ &Number of previous levels for context \\
        $L_d$ &Level from which context models are disabled\\
        $F$ &Dimension of feature vectors $\bm{\theta}$\\
        $\bm{C_p}$ &\textit{Context Fusor} to aggregate contexts \\
        $\bm{E_p}$ &\textit{Bit Estimator} to calculate bit consumption \\
        $K$ &Hash collision number of $\theta$ \\
        $k$ &A collided vertex out of $K$ \\
        $AOE$ &Area of effect of the vertex $n$ \\
        $PVF$ &Projected voxel feature for dimension-wise context of 3D to 2D \\
        $w$ &Normalized weights of vertices for hash fusion \\
        $L_{mse}$ &Mean Squared Error (MSE) loss, which measures fidelity \\
        $L_{entropy}$ &Entropy loss, which measures embedding size \\
        $\lambda$ &Tradeoff parameter to balance fidelity and size \\
        $\Omega$ &Parameter collection of the rendering MLP \\
        $\omega$ &A parameter of the collection $\Omega$ \\
        $\omega_q$ & The quantized parameter of $\omega$ \\
        $D$ & Number of digits for quantizing the rendering MLP \\
        $M$ & Number of $\theta$s in the embeddings \\
    \bottomrule[2pt]
    \end{tabular}
    \caption{Notation Table}
    \label{tab:notation_table}
\end{table*}


\begin{table*}
    \setlength\tabcolsep{7pt}
    \centering
    \begin{tabular}{lccccccccc}
    \toprule[2pt]
    Method & \textit{chair} & \textit{drums} & \textit{ficus} & \textit{hotdog} & \textit{lego} & \textit{materials} & \textit{mic} & \textit{ship} & \textit{Avg.} \\ \cline{1-10}
    \multicolumn{10}{c}{PSNR$\uparrow$}                    \\ \cline{1-10}
    Instant-NGP~\cite{INGP}     &35.91 &25.18 &33.76 &37.48 &35.86 &29.65 &36.98 &30.93 &33.22  \\
    SHACIRA~\cite{SHACIRA}     &31.88 &24.52 &30.65 &34.22 &31.79 &27.50 &32.00 &24.12 &29.59  \\
    MaskDWT($1e-10$)~\cite{MaskDWT}     &34.14 &25.53 &32.87 &35.93 &34.93 &29.54 &33.48 &29.15 &31.94  \\
    MaskDWT($5e-11$)~\cite{MaskDWT}     &34.52 &25.66 &33.03 &36.20 &35.16 &29.58 &33.68 &29.19 &32.13  \\
    MaskDWT($2.5e-11$)~\cite{MaskDWT}     &34.68 &25.56 &33.17 &36.37 &35.50 &29.56 &33.74 &29.34 &32.24  \\
    BiRF-F1~\cite{BiRF}     &33.38 &25.07 &32.26 &35.78 &33.52 &28.74 &34.42 &29.04 &31.53     \\
    BiRF-F2~\cite{BiRF}     &34.75 &25.59 &33.91 &36.59 &35.06 &29.49 &36.01 &29.74 &32.64     \\
    BiRF-F4~\cite{BiRF}     &35.66 &25.84 &34.42 &37.13 &36.02 &29.80 &36.91 &30.30 &33.26     \\
    BiRF-F8~\cite{BiRF}     &36.17 &26.05 &34.71 &37.51 &36.48 &30.09 &37.44 &30.27 &33.59     \\
    Ours($F=8, \lambda=4e-3$)     &34.76 &26.11 &34.15 &36.96 &35.38 &30.53 &36.64 &31.00 &33.19     \\
    Ours($F=8, \lambda=2e-3$)     &35.13 &26.08 &34.35 &37.28 &35.76 &30.63 &37.00 &31.46 &33.46     \\
    Ours($F=8, \lambda=1e-3$)     &35.37 &26.08 &34.46 &37.46 &35.98 &30.75 &37.31 &31.72 &33.64     \\
    Ours($F=8, \lambda=0.7e-3$)     &35.51 &26.18 &34.43 &37.42 &36.16 &30.72 &37.28 &31.83 &33.69     \\ \cline{1-10}
    \multicolumn{10}{c}{SSIM$\uparrow$}                    \\ \cline{1-10}
    Instant-NGP~\cite{INGP}     &0.986 &0.933 &0.983 &0.983 &0.981 &0.950 &0.992 &0.896 &0.963  \\
    SHACIRA~\cite{SHACIRA}     &0.967 &0.929 &0.969 &0.974 &0.966 &0.936 &0.980 &0.847 &0.946  \\
    BiRF-F1~\cite{BiRF}     &0.973 &0.921 &0.974 &0.973 &0.965 &0.934 &0.985 &0.877 &0.950  \\
    BiRF-F2~\cite{BiRF}     &0.980 &0.930 &0.981 &0.978 &0.976 &0.943 &0.989 &0.888 &0.958  \\
    BiRF-F4~\cite{BiRF}     &0.984 &0.934 &0.983 &0.980 &0.980 &0.948 &0.911 &0.895 &0.962  \\
    BiRF-F8~\cite{BiRF}     &0.986 &0.937 &0.984 &0.981 &0.982 &0.951 &0.992 &0.897 &0.964  \\
    Ours($F=8, \lambda=4e-3$)     &0.980 &0.941 &0.983 &0.978 &0.978 &0.958 &0.991 &0.901 &0.964  \\
    Ours($F=8, \lambda=2e-3$)     &0.982 &0.942 &0.984 &0.980 &0.980 &0.959 &0.992 &0.909 &0.966  \\
    Ours($F=8, \lambda=1e-3$)     &0.984 &0.941 &0.984 &0.982 &0.981 &0.960 &0.993 &0.913 &0.967  \\
    Ours($F=8, \lambda=0.7e-3$)     &0.984 &0.942 &0.984 &0.982 &0.982 &0.960 &0.993 &0.915 &0.968  \\ \cline{1-10}
    \multicolumn{10}{c}{LPIPS$\downarrow$}                    \\ \cline{1-10}
    Instant-NGP~\cite{INGP}     &0.021 &0.092 &0.024 &0.034 &0.022 &0.069 &0.014 &0.138 &0.052  \\
    SHACIRA~\cite{SHACIRA}     &0.045 &0.090 &0.043 &0.049 &0.045 &0.083 &0.032 &0.203 &0.074  \\
    BiRF-F1~\cite{BiRF}     &0.037 &0.086 &0.034 &0.045 &0.043 &0.078 &0.022 &0.141 &0.061  \\
    BiRF-F2~\cite{BiRF}     &0.024 &0.073 &0.024 &0.036 &0.025 &0.064 &0.016 &0.127 &0.049  \\
    BiRF-F4~\cite{BiRF}     &0.019 &0.066 &0.020 &0.032 &0.017 &0.057 &0.012 &0.117 &0.043  \\
    BiRF-F8~\cite{BiRF}     &0.016 &0.063 &0.018 &0.028 &0.015 &0.051 &0.009 &0.112 &0.039  \\
    Ours($F=8, \lambda=4e-3$)     &0.028 &0.071 &0.023 &0.043 &0.027 &0.057 &0.015 &0.140 &0.050  \\
    Ours($F=8, \lambda=2e-3$)     &0.024 &0.070 &0.022 &0.038 &0.024 &0.055 &0.012 &0.130 &0.047  \\
    Ours($F=8, \lambda=1e-3$)     &0.022 &0.071 &0.020 &0.035 &0.022 &0.054 &0.011 &0.124 &0.045  \\
    Ours($F=8, \lambda=0.7e-3$)     &0.021 &0.069 &0.020 &0.034 &0.021 &0.053 &0.011 &0.121 &0.044  \\ \cline{1-10}
    \multicolumn{10}{c}{SIZE(MB)$\downarrow$}                    \\ \cline{1-10}
    Instant-NGP~\cite{INGP}     &45.56 &45.56 &45.56 &45.56 &45.56 &45.56 &45.56 &45.56 &45.56  \\
    SHACIRA~\cite{SHACIRA}     &1.477 &1.527 &1.329 &1.739 &1.820 &1.766 &1.174 &2.162 &1.624  \\
    MaskDWT($1e-10$)~\cite{MaskDWT}     &0.985 &0.988 &1.011 &0.529 &0.787 &0.988 &0.555 &0.766 &0.826  \\
    MaskDWT($5e-11$)~\cite{MaskDWT}     &1.384 &1.404 &1.394 &0.750 &1.114 &1.401 &0.759 &1.090 &1.162  \\
    MaskDWT($2.5e-11$)~\cite{MaskDWT}     &1.988 &1.858 &1.968 &1.118 &1.647 &2.000 &1.208 &1.712 &1.687  \\
    BiRF-F1~\cite{BiRF}     &0.7 &0.7 &0.8 &0.7 &0.7 &0.8 &0.6 &0.8 &0.7  \\
    BiRF-F2~\cite{BiRF}     &1.3 &1.5 &1.4 &1.4 &1.4 &1.4 &1.3 &1.5 &1.4  \\
    BiRF-F4~\cite{BiRF}     &2.7 &2.9 &2.8 &2.8 &2.8 &2.8 &2.7 &3.0 &2.8  \\
    BiRF-F8~\cite{BiRF}     &5.6 &5.7 &5.8 &5.8 &5.8 &5.7 &5.6 &6.0 &5.8  \\
    Ours($F=8, \lambda=4e-3$)     &0.406 &0.488 &0.365 &0.332 &0.377 &0.485 &0.332 &0.560 &0.418  \\
    Ours($F=8, \lambda=2e-3$)     &0.511 &0.649 &0.444 &0.367 &0.454 &0.610 &0.366 &0.717 &0.515  \\
    Ours($F=8, \lambda=1e-3$)     &0.618 &0.852 &0.534 &0.420 &0.554 &0.727 &0.442 &0.915 &0.633  \\
    Ours($F=8, \lambda=0.7e-3$)     &0.689 &1.003 &0.588 &0.470 &0.602 &0.851 &0.471 &1.106 &0.722  \\
    \bottomrule[2pt]
    \end{tabular}
    \caption{Detailed quantitative results of storage size against fidelity quality (PSNR, SSIM, LPIPS) of each scene on NeRF-Synthetic dataset. We focus on NeRF compression approaches, along with our base model Instant-NGP. For quantitative results of other approaches, please refer to BiRF~\cite{BiRF} paper, as we do not duplicate them here.}
\end{table*}

\begin{table*}
    \setlength\tabcolsep{7pt}
    \centering
    \begin{tabular}{lcccccc}
    \toprule[2pt]
    Method & \textit{Barn} & \textit{Caterpillar} & \textit{Family} & \textit{Ignatius} & \textit{Truck} & \textit{Avg.} \\ \cline{1-7}
    \multicolumn{7}{c}{PSNR$\uparrow$}                    \\ \cline{1-7}
    Instant-NGP~\cite{INGP}     &28.19 &25.94 &34.32 &28.17 &27.03 &28.73  \\
    MaskDWT($1e-10$)~\cite{MaskDWT}     &26.49 &25.50 &32.57 &28.06 &26.21 &27.77  \\
    BiRF-F1~\cite{BiRF}     &27.11 &25.48 &33.21 &27.71 &26.80 &28.06  \\
    BiRF-F2~\cite{BiRF}     &27.65 &25.87 &33.86 &27.78 &27.31 &28.49  \\
    BiRF-F4~\cite{BiRF}     &27.74 &25.97 &34.33 &27.92 &27.46 &28.68  \\
    BiRF-F8~\cite{BiRF}     &27.69 &26.00 &34.45 &27.92 &27.54 &28.72  \\
    Ours($F=8, \lambda=8e-3$)     &28.15 &26.22 &33.23 &27.91 &27.53 &28.61  \\
    Ours($F=8, \lambda=4e-3$)     &28.32 &26.18 &33.60 &28.08 &27.57 &28.75  \\
    Ours($F=8, \lambda=2e-3$)     &28.51 &26.36 &33.80 &28.02 &27.48 &28.83  \\
    Ours($F=8, \lambda=0.7e-3$)     &28.76 &26.44 &34.12 &27.93 &27.62 &28.97  \\ \cline{1-7}
    \multicolumn{7}{c}{SSIM$\uparrow$}                    \\ \cline{1-7}
    Instant-NGP~\cite{INGP}     &0.881 &0.915 &0.968 &0.948 &0.918 &0.926  \\
    BiRF-F1~\cite{BiRF}     &0.851 &0.894 &0.955 &0.940 &0.894 &0.907  \\
    BiRF-F2~\cite{BiRF}     &0.869 &0.904 &0.963 &0.944 &0.907 &0.917  \\
    BiRF-F4~\cite{BiRF}     &0.877 &0.909 &0.966 &0.946 &0.914 &0.922  \\
    BiRF-F8~\cite{BiRF}     &0.882 &0.910 &0.968 &0.947 &0.917 &0.925  \\
    Ours($F=8, \lambda=8e-3$)     &0.866 &0.911 &0.955 &0.941 &0.910 &0.917  \\
    Ours($F=8, \lambda=4e-3$)     &0.872 &0.914 &0.959 &0.944 &0.914 &0.921  \\
    Ours($F=8, \lambda=2e-3$)     &0.879 &0.917 &0.961 &0.946 &0.917 &0.924  \\
    Ours($F=8, \lambda=0.7e-3$)     &0.884 &0.920 &0.965 &0.947 &0.921 &0.927  \\ \cline{1-7}
    \multicolumn{7}{c}{LPIPS$\downarrow$}                    \\ \cline{1-7}
    Instant-NGP~\cite{INGP}     &0.233 &0.161 &0.057 &0.087 &0.151 &0.138  \\
    BiRF-F1~\cite{BiRF}     &0.223 &0.159 &0.063 &0.080 &0.159 &0.137  \\
    BiRF-F2~\cite{BiRF}     &0.198 &0.144 &0.052 &0.075 &0.139 &0.122  \\
    BiRF-F4~\cite{BiRF}     &0.187 &0.136 &0.046 &0.072 &0.128 &0.114  \\
    BiRF-F8~\cite{BiRF}     &0.180 &0.133 &0.043 &0.072 &0.121 &0.109  \\
    Ours($F=8, \lambda=8e-3$)     &0.243 &0.159 &0.081 &0.087 &0.154 &0.145  \\
    Ours($F=8, \lambda=4e-3$)     &0.234 &0.154 &0.075 &0.084 &0.147 &0.139  \\
    Ours($F=8, \lambda=2e-3$)     &0.222 &0.149 &0.071 &0.083 &0.143 &0.134  \\
    Ours($F=8, \lambda=0.7e-3$)     &0.212 &0.145 &0.065 &0.080 &0.139 &0.128  \\ \cline{1-7}
    \multicolumn{7}{c}{SIZE(MB)$\downarrow$}                    \\ \cline{1-7}
    Instant-NGP~\cite{INGP}     &45.56 &45.56 &45.56 &45.56 &45.56 &45.56  \\
    MaskDWT($1e-10$)~\cite{MaskDWT}     &0.886 &1.219 &0.666 &0.769 &1.038 &0.916  \\
    BiRF-F1~\cite{BiRF}     &0.8 &0.8 &0.8 &0.8 &0.8 &0.8  \\
    BiRF-F2~\cite{BiRF}     &1.6 &1.6 &1.5 &1.6 &1.6 &1.6  \\
    BiRF-F4~\cite{BiRF}     &3.1 &3.1 &2.9 &3.2 &3.1 &3.1  \\
    BiRF-F8~\cite{BiRF}     &6.1 &6.0 &5.8 &6.3 &6.0 &6.0  \\
    Ours($F=8, \lambda=8e-3$)     &0.546 &0.579 &0.384 &0.432 &0.511 &0.490  \\
    Ours($F=8, \lambda=4e-3$)     &0.726 &0.824 &0.455 &0.559 &0.708 &0.654  \\
    Ours($F=8, \lambda=2e-3$)     &0.976 &1.067 &0.543 &0.721 &0.992 &0.860  \\
    Ours($F=8, \lambda=0.7e-3$)     &1.465 &1.652 &0.710 &1.146 &1.539 &1.302  \\
    \bottomrule[2pt]
    \end{tabular}
    \caption{Detailed quantitative results of storage size against fidelity quality (PSNR, SSIM, LPIPS) of each scene on Tanks and Temples dataset. We focus on NeRF compression approaches, along with our base model Instant-NGP. For quantitative results of other approaches, please refer to BiRF~\cite{BiRF} paper, as we do not duplicate them here.}
\end{table*}

\begin{table*}
    \setlength\tabcolsep{7pt}
    \centering
    \begin{tabular}{lccccccccc}
    \toprule[2pt]
    Setting of $F$ & \textit{chair} & \textit{drums} & \textit{ficus} & \textit{hotdog} & \textit{lego} & \textit{materials} & \textit{mic} & \textit{ship} & \textit{Avg.} \\ \cline{1-10}
    \multicolumn{10}{c}{PSNR$\uparrow$}                    \\ \cline{1-10}
    $F=1$     &33.17 &25.31 &32.38 &36.38 &33.57 &29.79 &33.60 &29.86 &31.74  \\
    $F=2$     &34.41 &25.77 &33.57 &36.94 &35.17 &30.20 &35.65 &30.94 &32.83  \\
    $F=4$     &35.32 &25.99 &34.32 &37.43 &36.01 &30.62 &36.90 &31.74 &33.54  \\
    $F=8$     &35.64 &26.06 &34.52 &37.48 &36.49 &30.76 &37.25 &31.81 &33.75  \\ \cline{1-10}
    \multicolumn{10}{c}{SSIM$\uparrow$}                    \\ \cline{1-10}
    $F=1$     &0.973 &0.929 &0.976 &0.977 &0.968 &0.950 &0.984 &0.883 &0.955  \\
    $F=2$     &0.980 &0.936 &0.981 &0.980 &0.977 &0.955 &0.990 &0.902 &0.963  \\
    $F=4$     &0.984 &0.941 &0.984 &0.982 &0.982 &0.959 &0.992 &0.914 &0.967  \\
    $F=8$     &0.985 &0.942 &0.985 &0.983 &0.983 &0.961 &0.993 &0.914 &0.968  \\ \cline{1-10}
    \multicolumn{10}{c}{LPIPS$\downarrow$}                    \\ \cline{1-10}
    $F=1$     &0.045 &0.094 &0.037 &0.044 &0.048 &0.071 &0.027 &0.159 &0.066  \\
    $F=2$     &0.031 &0.080 &0.027 &0.037 &0.030 &0.063 &0.016 &0.135 &0.052  \\
    $F=4$     &0.022 &0.071 &0.021 &0.032 &0.021 &0.054 &0.011 &0.122 &0.044  \\
    $F=8$     &0.020 &0.070 &0.019 &0.032 &0.019 &0.052 &0.010 &0.121 &0.043  \\ \cline{1-10}
    \multicolumn{10}{c}{SIZE(MB)$\downarrow$}                    \\ \cline{1-10}
    $F=1$     &0.827 &0.816 &0.806 &0.922 &0.838 &0.822 &0.802 &0.901 &0.842  \\
    $F=2$     &1.445 &1.434 &1.425 &1.530 &1.456 &1.442 &1.421 &1.531 &1.460  \\
    $F=4$     &2.699 &2.697 &2.680 &2.820 &2.710 &2.697 &2.676 &2.773 &2.719  \\
    $F=8$     &5.210 &5.202 &5.191 &5.334 &5.222 &5.208 &5.187 &5.281 &5.229  \\
    \bottomrule[2pt]
    \end{tabular}
    \caption{Detailed quantitative results of upper bounds (\ie $\lambda=0$) of our CNC model on NeRF-Synthetic dataset. In this case, no entropy constraint is applied to the embeddings, thus their size is equal to the amount of $\theta$s as each parameter consumes 1 bit. The rendering MLP is not quantized but retained in float32. Context models are excluded.}
\end{table*}

\begin{table*}
    \setlength\tabcolsep{7pt}
    \centering
    \begin{tabular}{lcccccc}
    \toprule[2pt]
    Setting of $F$ & \textit{Barn} & \textit{Caterpillar} & \textit{Family} & \textit{Ignatius} & \textit{Truck} & \textit{Avg.} \\ \cline{1-7}
    \multicolumn{7}{c}{PSNR$\uparrow$}                    \\ \cline{1-7}
    $F=8$     &28.68 &26.37 &34.33 &27.91 &27.48 &28.95  \\ \cline{1-7}
    \multicolumn{7}{c}{SSIM$\uparrow$}                    \\ \cline{1-7}
    $F=8$     &0.886 &0.920 &0.968 &0.948 &0.921 &0.928  \\ \cline{1-7}
    \multicolumn{7}{c}{LPIPS$\downarrow$}                    \\ \cline{1-7}
    $F=8$     &0.209 &0.146 &0.059 &0.080 &0.138 &0.126  \\ \cline{1-7}
    \multicolumn{7}{c}{SIZE(MB)$\downarrow$}                    \\ \cline{1-7}
    $F=8$     &5.326 &5.277 &5.315 &5.362 &5.263 &5.309  \\
    \bottomrule[2pt]
    \end{tabular}
    \caption{Detailed quantitative results of upper bounds (\ie $\lambda=0$) of our CNC model on Tanks and Temples dataset. In this case, no entropy constraint is applied to the embeddings, thus their size is equal to the amount of $\theta$s as each parameter consumes 1 bit. The rendering MLP is not quantized but retained in float32. Context models are excluded.}
\end{table*}

\begin{figure*}
    \centering
    \includegraphics[width=0.9\linewidth]{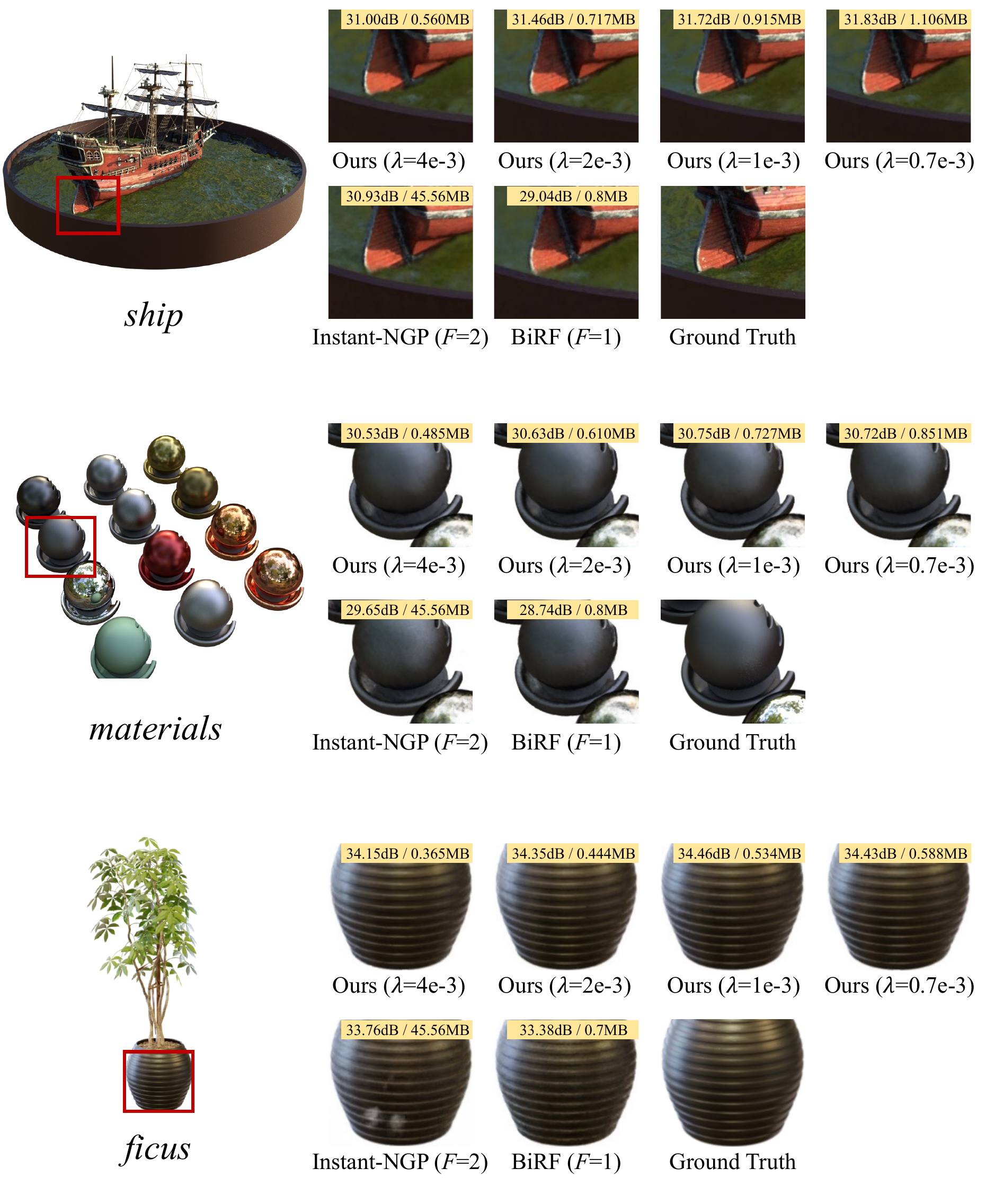}
    \caption{Qualitative quality comparison of Synthetic-NeRF dataset. Quantitative results of PSNR/size are shown in the upper right.}
    \label{fig:supple_NeRF}
\end{figure*}

\begin{figure*}
    \centering
    \includegraphics[width=0.9\linewidth]{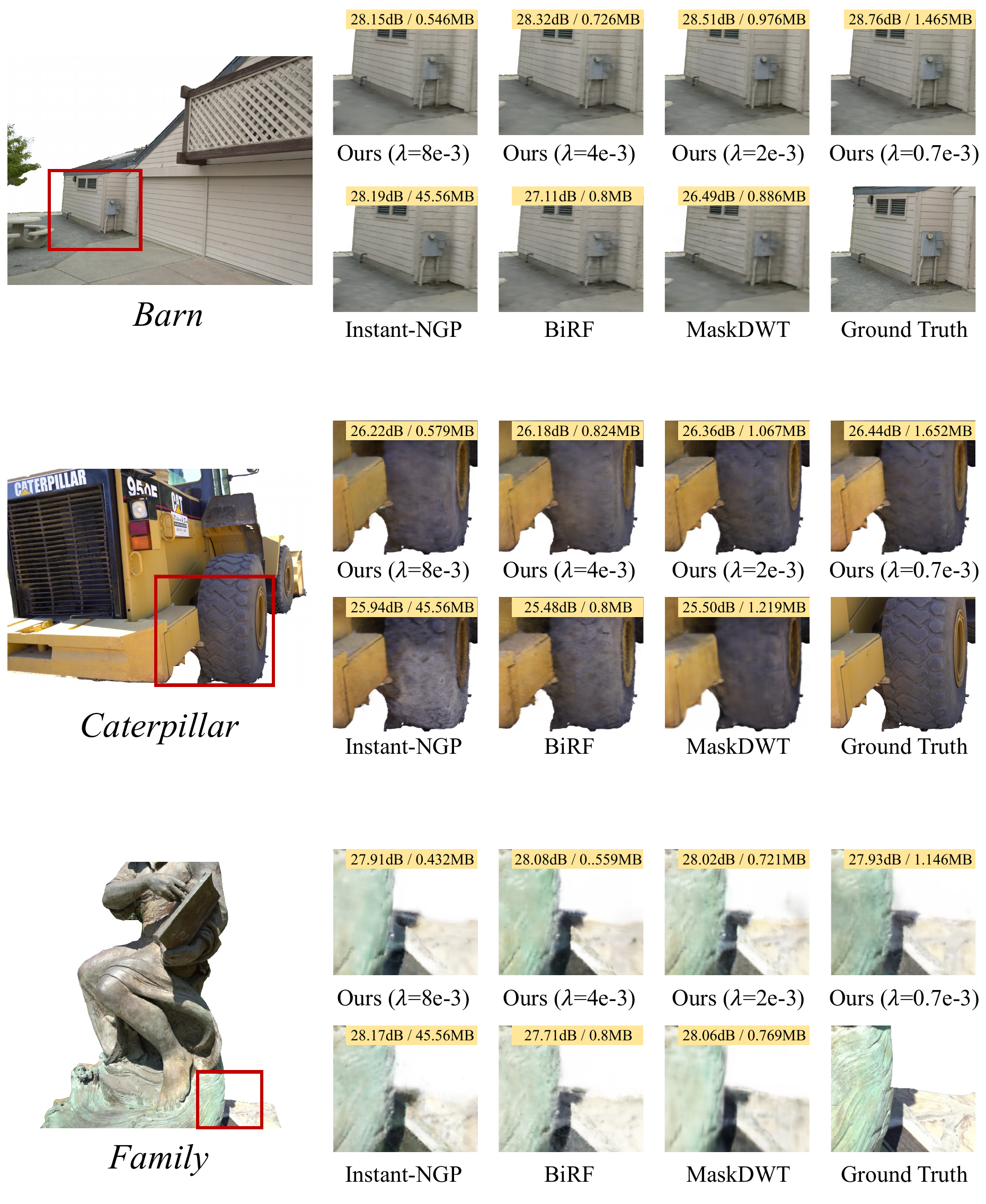}
    \caption{Qualitative quality comparison of Tanks and Temples dataset. Quantitative results of PSNR/size are shown in the upper right.}
    \label{fig:supple_Tanks}
\end{figure*}

\end{document}